%% file: main.tex
\documentclass[twoside]{article}

\usepackage[accepted]{aistats2020}
\usepackage[round,authoryear]{natbib}

\usepackage{subfig}
\usepackage{wrapfig}
\usepackage{graphicx}
\usepackage{lastpage}
\usepackage{amsmath}
\usepackage{amssymb}
\usepackage{multirow}
\usepackage[table,xcdraw]{xcolor}
\usepackage[shortlabels]{enumitem}
\usepackage[textsize=tiny]{todonotes}
\usepackage[title]{appendix}

\usepackage{amsthm}
\newtheorem{theorem}{Theorem}

\newtheorem{proposition}{Proposition}

\usepackage[frozencache,cachedir=.]{minted}
\setminted{fontsize=\small,baselinestretch=1}

\usepackage[T1]{fontenc}
\usepackage{inconsolata}

\usepackage{booktabs}
\usepackage{arydshln}

\newcommand*\rot{\rotatebox{90}}

\usepackage{Definitions}

\newcommand{\tp}{\text{ \texttt{TP} }}

\newcommand{\tn}{\text{ \texttt{TN} }}
\newcommand{\ap}{\text{ \texttt{AP} }}
\newcommand{\an}{\text{ \texttt{AN} }}
\newcommand{\pp}{\text{ \texttt{PP} }}
\newcommand{\pn}{\text{ \texttt{PN} }}
\newcommand{\all}{\text{ \texttt{ALL} }}

\newcommand{\tpz}{\text{\texttt{TP}}}
\newcommand{\fpz}{\text{\texttt{FP}}}
\newcommand{\fnz}{\text{\texttt{FN}}}
\newcommand{\tnz}{\text{\texttt{TN}}}
\newcommand{\apz}{\text{\texttt{AP}}}
\newcommand{\anz}{\text{\texttt{AN}}}
\newcommand{\ppz}{\text{\texttt{PP}}}
\newcommand{\pnz}{\text{\texttt{PN}}}
\newcommand{\allz}{\text{\texttt{ALL}}}

\usepackage[colorlinks = true,
            linkcolor = black,
            urlcolor  = blue,
            citecolor = black,
            anchorcolor = black]{hyperref}

\begin{document}

%

%

\twocolumn[

\aistatstitle{AP-Perf: Incorporating Generic Performance Metrics in Differentiable Learning}

\aistatsauthor{ Rizal Fathony \And J. Zico Kolter }

\aistatsaddress{ Carnegie Mellon University \\rfathony@cs.cmu.edu \And  Carnegie Mellon University and Bosch Center for AI \\zkolter@cs.cmu.edu } ]

\begin{abstract}
  We propose a method that enables practitioners to conveniently incorporate custom non-decomposable performance metrics into differentiable learning pipelines, notably those based upon neural network architectures. Our approach is based on the recently developed adversarial prediction framework, a distributionally robust approach that optimizes a metric in the worst case given the statistical summary of the empirical distribution. We formulate a marginal distribution technique to reduce the complexity of optimizing the adversarial prediction formulation over a vast range of non-decomposable metrics. We demonstrate how easy it is to write and incorporate complex custom metrics using our provided tool. 
  Finally, we show the effectiveness of our approach various classification tasks on tabular datasets from the UCI repository and benchmark datasets, as well as image classification tasks.  
  The code for our proposed method is available at \href{https://github.com/rizalzaf/AdversarialPrediction.jl}{\small \texttt{https://github.com/\linebreak rizalzaf/AdversarialPrediction.jl}}.
\end{abstract}

\section{INTRODUCTION}

In real-world applications, the performance of machine learning algorithms is measured with evaluation metrics specifically tailored to the problem of interest.  Although the accuracy is the most popular evaluation metric, many applications require the use of more complex evaluation metrics that are not additively decomposable into sample-wise measures.  For example, in text classification area, F$_\beta$  score (weighted harmonic mean of precision and recall) is frequently used to evaluate the  performance. F$_\beta$  is also popular in the classification tasks with imbalanced datasets. In medical fields, the sensitivity, specificity, and informedness are some of the popular evaluation metrics.  Many of these performance metrics require inherent trade-offs, for example, balancing precision versus recall.

A variety of learning algorithms that incorporate some of the performance metrics above into their learning objectives have been proposed. One of the first approaches to this problem is the SVM-Perf \citep{joachims2005support}, which augments the constraints of a binary SVM optimization with the metrics. \citet{koyejo2014consistent} and \citet{narasimhan2014statistical} proposed plug-in classifiers that rely on an external estimator of class probability (typically using logistic regression).
\citet{hazan2010direct} proposed a way to directly optimizes the performance metric by computing the asymptotic gradient of the metric.
Some of the previous research focused only on a specific performance metric, most notably, the F1-score \citep{dembczynski2011exact,parambath2014optimizing,lipton2014optimal,wang2015adversarial,shi2017bregman}. Optimizing the metric on specific learning settings have also been explored, for example, in online learning \citep{busa2015online,kar2014online,narasimhan2015optimizing} and ranking \citep{yue2007support,narasimhan2013structural,narasimhan2013svm,kar2015surrogate}. Finally, several efforts have been made to incorporate non-decomposable metrics into neural networks training \citep{eban2017scalable,song2016training,sanyal2018optimizing}.

\begin{figure*}[t!]
\begin{minipage}{0.38\textwidth}
\begin{minted}[frame=lines,
               framesep=2mm]{text}
model = Chain(
    Dense(nvar, 100, relu),
    Dense(100, 100, relu),
    Dense(100, 1), 
    vec)       

objective(x, y) = mean(
  logitbinarycrossentropy(model(x), y))

opt = ADAM(1e-3)  
Flux.train!(objective, params(model), 
  train_set, opt)
\end{minted}
\end{minipage}%
\begin{minipage}{0.03\textwidth}
$\phantom{a}$
\end{minipage}
\begin{minipage}{0.58\textwidth}
\begin{minted}[frame=lines,
               framesep=2mm]{text}
model = Chain(Dense(nvar, 100, relu),
    Dense(100, 100, relu), Dense(100, 1), vec)     

@metric FBeta beta
function define(::Type{FBeta}, C::ConfusionMatrix, beta)
    return ((1 + beta^2) * C.tp) / (beta^2 * C.ap + C.pp)  
end   
f2_score = FBeta(2)
special_case_positive!(f2_score)

objective(x, y) = ap_objective(model(x), y, f2_score)
Flux.train!(objective, params(model), train_set, ADAM(1e-3))
\end{minted}
\end{minipage}
\caption{Code examples for incorporating F$_2$-score metric into a neural network training pipeline (right), compared with the standard code for cross-entropy objective (left). The codes are implemented in Julia.}
\label{fig:code}
\end{figure*}

Despite this rich literature on learning algorithms for non-decomposable metrics, the algorithms have not been widely used in practical applications, particularly in the modern machine learning applications that rely on the representational power of neural network architectures, where training is typically done using a gradient-based method.  Instead of being trained to optimize the evaluation metric of interest, they are typically trained to minimize cross-entropy loss, with the hope that it will indirectly optimize the metric \citep{eban2017scalable}. However, as mentioned in previous research \citep{cortes2004auc,davis2006relationship}, this discrepancy between the target  and optimized metric may lead to inferior results.

We argue that two factors hinder the wide adoption of the learning algorithms for non-decomposable metrics into many modern machine learning applications. First, many of the existing learning algorithms are not flexible enough to accommodate the custom need of real-world applications. Their formulations only cover a few types of performance metrics that may not be relevant for some applications. Second, even though some of the existing formulations are flexible, they do not provide a way for practitioners to customize the usage. The authors of these flexible methods oftentimes only provide few uses case metrics in their experiments and also their published codes. A significant amount of effort (e.g., deriving the formulations and rewriting the codes) need to be spent by a practitioner who wants to implement and customize their algorithm to the specific needs of the applications. This also still be a problem even for the latest development of algorithms that already specifically target neural network training. 
These two factors force many practitioners to choose a method that is easy to incorporate to their machine learning system, for example, the cross-entropy objective (a common proxy for accuracy metric) that is readily available in many  frameworks. 

In this paper, our goal is to overcome the problem above. We propose a 
generic framework for optimizing arbitrary complex non-decomposable performance metrics using gradient-based learning procedures.  Our framework can be applied to most of the common use-cases of non-decomposable metrics. 
Specifically, we require the metric to be derived from the value of the confusion matrix with minimal requirements on how the metric needs to be constructed. Our formulation also supports optimizing a performance metric with a constraint over another metric. This is useful in the case where we want to balance the trade-off between two metrics, for example, in the case where we want to maximize precision subject to recall $\ge 0.8$.  Our approach is based on the adversarial prediction framework \citep{fathony2018consistent,asif2015adversarial}, a distributionally robust framework for constructing learning algorithms that seeks a predictor that maximizes the performance metric in the worst case given the statistical summary of the empirical distribution. We replace the empirical data for evaluating the predictor with an adversary that is free to choose an evaluating distribution from the set of conditional distribution that matches the statistics of empirical data via moment matching on the features.  Although naively applying this approach is not possible, we develop a marginalization technique that reduces the number of variables in the resulting optimization from exponentially many variables to just quadratic.

In addition to these algorithmic contributions, we establish the Fisher consistency of the method, a feature notably lacking from much past work approximately optimizing performance metrics \citep{tewari2007consistency,liu2007fisher}. We also develop a programming interface such that a practitioner can easily construct the metric and integrate it into their learning pipeline. Figure \ref{fig:code} provides an example of incorporating the F$_2$-score metric into the training pipeline of our method. Notice that only minimal changes from the standard cross-entropy learning code are needed.
Finally, we evaluate the performance of our method against the standard training on several benchmark datasets within neural network learning pipelines and demonstrate that our method vastly outperforms traditional approaches for training these networks.

\section{BACKGROUND}

\subsection{Performance Metrics}

\begin{table}[tb]
\centering
\small
\caption{Confusion Matrix}
\label{tbl:confusion-matrix}
\begin{tabular}{c|c|c|c|c}
\cline{3-4}
\multicolumn{2}{c|}{}                                                                                        & \multicolumn{2}{c|}{Actual}                                                                                                                                                 &                                                                                                                 \\ \cline{3-4}
\multicolumn{2}{c|}{\multirow{-2}{*}{}}                                                                      & Positive                                                                                & Negative                                                                                & \multirow{-2}{*}{}                                                                                              \\ \hline
\multicolumn{1}{|c|}{}                                                                            & Positive & \begin{tabular}[c]{@{}c@{}}True \\ Pos. (TP)\end{tabular}                           & \begin{tabular}[c]{@{}c@{}}False \\ Pos. (FP)\end{tabular}                          & \multicolumn{1}{c|}{\cellcolor[HTML]{EFEFEF}\begin{tabular}[c]{@{}c@{}}Predicted \\ Pos. (PP)\end{tabular}} \\ \cline{2-5} 
\multicolumn{1}{|c|}{\multirow{-2}{*}{\vspace{-5pt}\rot{Pred.}}} & Negative & \begin{tabular}[c]{@{}c@{}}False \\ Neg. (FN)\end{tabular}                          & \begin{tabular}[c]{@{}c@{}}True \\ Neg. (TN)\end{tabular}                           & \multicolumn{1}{c|}{\cellcolor[HTML]{EFEFEF}\begin{tabular}[c]{@{}c@{}}Predicted \\ Neg. (PN)\end{tabular}} \\ \hline
\multicolumn{2}{c|}{}                                                                                        & \cellcolor[HTML]{EFEFEF}\begin{tabular}[c]{@{}c@{}}Actual \\ Pos. (AP)\end{tabular} & \cellcolor[HTML]{EFEFEF}\begin{tabular}[c]{@{}c@{}}Actual \\ Neg. (AN)\end{tabular} & \multicolumn{1}{c|}{\cellcolor[HTML]{C0C0C0}
\begin{tabular}[c]{@{}c@{}}All Data \\ (ALL)\end{tabular}}                                                     \\ \cline{3-5} 
\end{tabular}
\end{table}

Deciding on what performance metric to be used for evaluating the prediction is an important aspect of machine learning applications, since it will also guide the design of learning algorithms. A performance metric should be carefully picked to reflect the objective goal of the prediction \citep{powers2011evaluation}. Different tasks in machine learning  require different metrics that align  well with the tasks. For binary classification problems, many of the commonly used performance metrics are derived from the confusion matrix. 
The confusion matrix is a table that reports the values that relate the prediction of a classifier with the ground truth labels. Table \ref{tbl:confusion-matrix} shows the anatomy of the confusion matrix.

Most commonly used performance metrics can be derived from the confusion matrix. Some of the metrics are decomposable, which means that it can be broken down to an independent sum of another metric that depends only on a single sample. However, most of the interesting performance metrics are non-decomposable, where we need to consider all samples at once. There is a wide variety of non-decomposable performance metrics. Table \ref{tbl:perf-metric} shows some of the popular metrics and the formula on how to derive the metric from the confusion matrix.

\begin{table}[b]
\caption{Examples of Non-Decomposable Performance Metrics} \label{tbl:perf-metric}
\begin{center}
\begin{tabular}{ll}
\textbf{NAME}  &\textbf{FORMULA} \\
\hline \\
F$_\beta$-score             & $\frac{(1 + \beta^2) \tp}{\beta^2 \ap + \pp}$ \\ \addlinespace[2pt]
Geom. mean of Prec. \& Recall & $\frac{\tp}{\sqrt{\pp \cdot \ap}}$ \\ \addlinespace[2pt]
Balanced Accuracy         &  $ \frac{1}{2}\left(\frac{\tp}{\ap} + \frac{\tn}{\an} \right) $ \\ \addlinespace[2pt]
Bookmaker Informedness & $\frac{\tp}{\ap} + \frac{\tn}{\an} - 1$ \\ \addlinespace[2pt]
Cohen's kappa score & \\
\multicolumn{2}{r}{$\qquad\qquad\frac{  \left( \tp + \tn \right) / \all - \; \left( \ap \cdot \pp + \an \cdot \pn \right) / \all^2 } { 1 - \left( \ap \cdot \pp + \an \cdot \pn \right) / \all^2 }$} \\
\addlinespace[2pt]
Matthews correlation coefficient & \\
\multicolumn{2}{r}{$\qquad\qquad\frac{  \tp  / \all - \; \left( \ap \cdot \pp  \right) / \all^2 } { \sqrt{ \ap \cdot \pp \cdot \an \cdot \pn } / \all^2 }$} \\
\hline
\end{tabular}
\end{center}
\end{table}

\subsection{Existing Methods}

Many existing methods have been proposed for optimizing non-decomposable metrics. However, 
they do not facilitate an easy way to implement the methods on new custom tasks. They also do not provide convenient ways to integrate the algorithms to differentiable learning pipeline on custom non-decomposable performance metrics. 
SVM-Perf \citep{joachims2005support} is a large margin technique that enables the incorporation of a performance metric to the SVM training objective. However, for new metrics that are not explained in the paper, we need to formulate and implement an algorithm to find the maximum violated constraints for the given metric inside its cutting plane algorithm. 
Plug-in based classifiers \citep{koyejo2014consistent,narasimhan2014statistical} need to first solve probability estimation problems optimally, and then tune a threshold depending on the performance metric they optimize. This makes the techniques hard to incorporate into differentiable learning pipelines. Many existing methods only focus on developing formulations for specific performance metrics or providing examples on a few metrics without any complete guide on extending the methods to other metrics  \citep{hazan2010direct,dembczynski2011exact,parambath2014optimizing,lipton2014optimal,busa2015online,wang2015adversarial,shi2017bregman}.
Finally, even though some of the existing methods \citep{eban2017scalable,song2016training,sanyal2018optimizing} specifically targeted their approach to neural network learning, they do not provide an easy way to implement their method on new custom  metrics.

\subsection{Adversarial Prediction}

Recently developed adversarial prediction framework \citep{fathony2018consistent,asif2015adversarial} provides an alternative to the empirical risk minimization framework (ERM) \citep{vapnik1992principles} for designing learning algorithms. 
In a classification setting, the ERM framework prescribes the use of convex surrogate loss function as a tractable approximation to the original non-convex and non-continuous objective of optimizing an evaluation metric (e.g., accuracy).
In contrast, the adversarial prediction framework replaces the empirical training data for evaluating the metric with an adversary that is free to choose an evaluating distribution that approximates the training data. This approximation of the training data is performed by constraining the adversary's distribution to match the feature statistics of the empirical training data. 
Even though we started with a non-convex and non-continuous metric, 
the resulting optimization objective  is always convex with respect to the optimized variable.

The adversarial prediction framework has been previously used to design learning algorithms for many decomposable metrics, including the zero-one loss \citep{fathony2016adversarial}, ordinal regression loss \citep{fathony2017adversarial}, abstention loss \citep{fathony2018consistent}, cost-sensitive loss metrics \citep{asif2015adversarial}.
The extensions of the framework to  non-decomposable metrics and structured prediction have also been explored. The main challenge of these extensions is that naively solve the resulting dual optimization is intractable since we have to simultaneously consider all possible label assignments for all samples in the dataset. Previous research have tried to reduce the complexity of solving the problem.
One of the first efforts by \citet{wang2015adversarial} uses a double oracle technique to solve the problem for a few performance metrics (F$_1$-score, precision@k, and DCG). However, the double oracle algorithm they use does not have any guarantee that it will converge to the solution in polynomial time. Additionally, extending the approach to other metrics is hard since we have to formulate an algorithm to find the best response for the given metric, which is harder than the SVM-Perf's problem on finding the most violated constraint.

The second wave of research have been proposed for applying the adversarial prediction to non-decomposable metrics and structured prediction using marginalization technique that reduces the optimization over full exponentially sized conditional distributions into their polynomially sized marginal distributions. This technique has been applied to the problem of optimizing the F$_1$-score metric \citep{shi2017bregman}, tree-structured graphical models \citep{fathony2018distributionally}, and bipartite matching in graphs \citep{fathony2018efficient}.
However, these methods only focus on the specific performance metrics, and they do not provide a way to extend the method to custom performance metrics easily. 
Our paper is the first effort to generalize the marginalization technique to a vast range of performance metrics. Our approach is also the first method that can be easily integrated into differentiable learning pipelines.

\section{APPROACH}

To achieve our goal of providing a flexible and easy to use method for optimizing custom performance metrics, we formulate it as an adversarial prediction task. 

\subsection{Adversarial Prediction Formulation}

In a binary classification task, the training examples consist of pairs of training data and labels $\{({\bf x}_1, y_{1}), \hdots , ({\bf x}_n, y_{n}) \}$ drawn i.i.d  from a distribution $D$ on $\mathcal{X} \times \mathcal{Y}$, where $\mathcal{X}$ is the feature space and $\mathcal{Y} = \{0,1\}^n$ is the set of binary labels. A classifier needs to make a prediction $\yhat_i$  for each sample $\xvec_i$. The prediction is evaluated using a non-decomposable performance metric, $\text{metric}(\yvechat, \yvec)$. Here, we need to consider the prediction for all samples (denoted in vector notations) to compute the metric. 

The adversarial prediction method seeks a predictor that robustly maximizes the performance metric against an adversary that is constrained to approximate the training data (via moment matching constraints on the features) but otherwise aims to minimize the metric. Both predictor and adversary players are allowed to make probabilistic predictions over all possible label outcomes. 
Denote $\Pcal(\hat{\Yvec}) \triangleq \Phat(\hat{\Yvec}|{\bf X})$ as the predictor's probabilistic prediction and $\Qcal(\check{\Yvec}) \triangleq \Pchk(\check{\Yvec}|{\bf X})$ as the adversary's distribution.\footnote{Lowercase $y$ and  $\yvec$, denote scalar and vector values, and capitals, $Y$ or $\Ybf$, denote
random variables.} The adversary player needs to approximate the training data by selecting a conditional probability $\Qcal(\check{\Yvec})$ whose feature expectations match the empirical feature statistics. On the other hand, the predictor is free to choose any conditional probability $\Pcal(\hat{\Yvec}) $ that maximizes the expected metric. Formally, the adversarial prediction is formulated as: 
\begin{align}
    &\max_{\Pcal(\hat{\Yvec}) } \; \min_{\Qcal(\check{\Yvec}) 
    } \; 
    \mathbb{E}_{\tilde{P}(\Xbf); \Pcal(\hat{\Yvec});
    	\Qcal(\check{\Yvec})} \left[\text{metric}(\hat{\Yvec}, \check{\Yvec}) \right] \nonumber \\ 
    &\text{s.t.: } \mathbb{E}_{\tilde{P}(\Xbf);\Qcal(\check{\Yvec})}[\phi({\bf X},\check{\Yvec})]
    = \mathbb{E}_{\tilde{P}(\Xbf, \Ybf)}\left[\phi({\bf X},{\Yvec}) \right], \label{eq:primal}
\end{align}
where $\tilde{P}$ denotes the empirical distribution.
Using the method of Lagrangian multipliers and strong duality for convex-concave saddle
point problems \citep{von1945theory,sion1958general}, the dual formulation of Eq. \eqref{eq:primal} can be written as:
\begin{align}
	&\max_{\theta}
     \mathbb{E}_{\tilde{P}(\Xbf, \Ybf)} \bigg[
     \min_{\Qcal(\check{\Yvec})}
     \max_{\Pcal(\hat{\Yvec}) }
    \mathbb{E}_{\Pcal(\hat{\Yvec});
    	\Qcal(\check{\Yvec})}\Big[ \text{metric}(\hat{\Yvec},\check{\Yvec}) \notag \\
    & \qquad \qquad \qquad - \theta^\intercal \left( \phi({\bf X},\check{\Yvec}) - \phi({\bf X},{\Yvec}) \right)
    \Big]
    \bigg], \label{eq:dual}
\end{align}
where $\theta$ is the Lagrange dual variable for the moment matching constraints of the adversary's distribution. 
This follows directly from previous results in adversarial prediction \citep{fathony2018consistent}.

\subsection{Adversarial Prediction for Non- Decomposable Performance Metrics}
\label{sec:formulation}

We consider a family of performance metrics that can be expressed as a sum of fractions of the entities in the confusion matrix (Table \ref{tbl:confusion-matrix}):
\begin{equation}
    \text{metric}(\yvechat,\yvec) = \sum_j \frac{ a_j \tpz + b_j \tnz + f_j(\ppz,\apz)}{g_j(\ppz, \apz)}, \label{eq:metric}
\end{equation}
where $a_j$ and $b_j$ are constants, whereas $f_j$ and $g_j$ are  functions over $\ppz$ and $\apz$.
Hence, the numerator is a linear function over true positive ($\tpz$) and true negative ($\tnz$) which may also depends on sum statistics, i.e., predicted and actual positive ($\ppz$ and $\apz$) as well as their negative counterparts (predicted and actual negative ($\pnz$ and $\anz$)) and all data ($\allz$)\footnote{We simplify the inputs of $f_j$ and $g_j$ to be just $\ppz$ and $\apz$ since the other terms can be derived from $\ppz$ and $\apz$. $\allz$ is just a constant, whereas $\pnz = \allz - \ppz$ and $\anz = \allz - \apz$.}.
The denominator depends only on the sum statistics. 
This  metric construction 
covers a vast range of commonly used metrics, including all metrics in Table \ref{tbl:perf-metric}.

Applying the adversarial prediction framework to classification problems with non-decomposable metrics is non-trivial.
We take a look at the inner minimax problem of the dual formulation (Eq. \eqref{eq:dual}), i.e.:
\begin{equation}
	\min_{\Qcal(\check{\Yvec})} 
     \max_{\Pcal(\hat{\Yvec}) } 
    \mathbb{E}_{\Pcal(\hat{\Yvec});
    	\Qcal(\check{\Yvec})} \left[ \text{metric}(\hat{\Yvec},\check{\Yvec}) - \theta^\intercal \phi({\bf X},\check{\Yvec})  \right]. \label{eq:inner-minimax}
\end{equation}
Note that we set aside the empirical potential term ($\theta^\intercal \phi({\bf X},\Yvec) $) since it does not influence the inner minimax solution.
Unlike many previous adversarial prediction research \citep{asif2015adversarial,fathony2016adversarial,fathony2017adversarial,fathony2018consistent}, we cannot reduce the problem to sample-wise minimax problems since our metric is now non-decomposable. We need to deal with the full conditional distribution ($\Pcal(\yvechat)$ and $\Qcal(\yvechat)$) over all samples which is exponential in size. Therefore, naively solving the inner minimax problem is intractable.
In the subsequent analyses, we aim to reduce the complexity of solving the problem by optimizing over the  marginal distribution of $\Pcal(\yvechat)$ and $\Qcal(\yvechat)$.

We take a look at the expectation of the metric. 
We now define the marginal probability of the event where $y_i = 1$ and $\sum_{i'} y_{i'} = k$, which we write as $\Pcal(\yhat_i=1, {\textstyle\sum_{i'}} \yhat_{i'} = k))$ and $\Qcal(\ychk_i=1, {\textstyle\sum_{i'}} \ychk_{i'} = k))$ for the predictor and adversary respectively. Similarly, we also define the marginal probability of the event where $y_i = 0$ and $\sum_{i'} y_{i'} = k$.
Let us denote $\pvec_k^a$ be a vector with $n$ items where each of its items $(\pvec_k^a)_i$ represents the predictor's marginal probability $\Pcal(\yhat_i=a, {\textstyle\sum_{i'}} \yhat_{i'} = k))$. Similarly, we also denote $\qvec_l^a$ for the adversary's marginal probabilities. 
We also denote the marginal probability of sums as $r_k = \Pcal(\sum_i \yhat_i=k)$, and $s_l = \Qcal(\sum_i \ychk_i=l)$
Using these notations, we simplify the computation of the expected value of the performance metric in terms of  these marginal probabilities as stated in Theorem \ref{thm:marginal}.\footnote{The proof of this theorem and others in the paper are contained in Appendix \ref{sec:proof-appendix}.}

\begin{theorem}
\label{thm:marginal}
Given a performance metric that follows the construction in Eq. \eqref{eq:metric}, the expected value of the metric over exponentially sizes conditional probabilities $\Pcal(\hat{\Yvec}) $ and $\Qcal(\check{\Yvec})$ can be expressed as the sum of functions over marginal probability variables $\pvec_k^1$, $\qvec_l^1$, $\pvec_k^0$, $\qvec_l^0$, $r_k$, and $s_l$ as follows:
\begin{align}
    &\mathbb{E}_{\Pcal(\hat{\Yvec});
    	\Qcal(\check{\Yvec})}\left[ \text{metric}(\hat{\Yvec},\check{\Yvec}) \right] \!= \!\! \sum_{k \in [0,n]} \sum_{l \in [0,n]} \sum_j \frac{1}{g_j(k, l)} \big\{ \notag \\
    & \qquad \quad a_j [ \pvec_k^1 \cdot \qvec_l^1] + b_j [ \pvec_k^0 \cdot \qvec_l^0] + f_j(k,l) r_k s_l \big\}.  \label{eq:marginal-0-to-n-vec}
\end{align}
\end{theorem}

Some performance metrics (e.g. precision, recall, F-score, sensitivity, and specificity) enforce special cases to avoid division by zero. For the metrics that contains true positive, the special cases is usually defined as:
\begin{align}
    &\text{metric}(\zerovec,\zerovec) = 1; \quad
    \text{metric}(\zerovec,\yvec) = 0, \forall \yvec \neq \zerovec; \label{eq:special-case-pos}\\
    &\text{metric}(\yvechat,\zerovec) = 0, \forall \yvechat \neq \zerovec, \notag
\end{align}
whereas for the one with true negative, their cases are:
\begin{align}
    &\text{metric}(\onevec,\onevec) = 1; \quad
    \text{metric}(\onevec,\yvec) = 0, \forall \yvec \neq \onevec; \label{eq:special-case-neg}\\
    &\text{metric}(\yvechat,\onevec) = 0, \forall \yvechat \neq \onevec. 
    \notag
\end{align}
Here $\yvechat = \zerovec$ and  $\yvechat = \onevec$ means that the classifier predicts all samples as negative and positive respectively.
If the special cases are enforced, we need to modify Eq. \eqref{eq:marginal-0-to-n-vec} accordingly. For example, if both special cases for true positive and true negative are enforced, it becomes:
\begin{align}
 &\sum_{k \in [1,n\!-\!1]} \sum_{l \in [1,n\!-\!1]} \sum_j  \frac{1}{g_j(k, l)} \big\{ a_j [ \pvec_k^1 \cdot \qvec_l^1] + b_j [ \pvec_k^0 \cdot \qvec_l^0]  \notag \\ 
 & \qquad  + f_j(k,l) r_k s_l \big\} +  \Pcal(\zerovec) \Qcal(\zerovec) + \Pcal(\onevec) \Qcal(\onevec). \label{eq:marginal-vec-special-case-pos-neg}
\end{align}

Let us denote a $n \times n$ marginal distribution matrix $\Pbf$ where each column $\Pbf_{(:,k)}$ represents $\pvec^1_k$. Similarly, we denote a matrix $\Qbf$ for $\qvec^1_k$.
For our feature, we use additive feature function, i.e., $\phi(\xvec, \yvec) =\sum_i \phi(\xvec_i, y_i)$. For simplicity, we also define $\phi(\xvec_i, y_i = 0) =0$. Let us denote $\Psi$ be a $n \times m$ matrix where each of its columns denotes the feature for each sample, i.e.,  $\Psi_{:,i} = \phi(\xvec_i, y_i =1)$, and $m$ is the number of features. 
Using these notations, we simplify the dual formulation of the adversarial prediction in Theorem \ref{thm:theta_q_p}.

\begin{theorem}
\label{thm:theta_q_p}
Let $\Pbf$ and $\Qbf$ be the marginal predictor and adversary probability matrices respectively. Given a performance metric that follows the construction in Eq. \eqref{eq:metric} and features that are additive over each sample, the dual optimization formulation (Eq. \eqref{eq:dual}) can be equivalently computed as:
\begin{align}
	&\max_{\theta} \bigg\{ 
     \min_{\Qbf \in \Delta}
     \max_{\Pbf \in \Delta}  \bigg[
      \sum_{k,l \in [0,n]} \sum_j \tfrac{1}{g_j(k, l)} \Big\{ a_j [ \pvec_k^1 \cdot \qvec_l^1] \label{eq:dual-marginal} \\
    &  \!+\! b_j [ \pvec_k^0 \cdot \qvec_l^0]  \!+\! f_j(k,l) r_k s_l \Big\} \! - \! \langle \Qbf^\intercal \onevec, \Psi^\intercal \theta \rangle \bigg] \!+\!  \langle \yvec, \Psi^\intercal \theta \rangle \!\bigg\}\!, \notag 
\end{align}
where $\Delta$ is the set of valid marginal probability matrices denoted as:
\begin{align}
    \Delta = \left\{\Pvec \middle\vert 
    \begin{matrix}
    p_{i,k} \geq 0 &\quad \forall i,k \in [1,n] \\
    p_{i,k} \leq \tfrac{1}{k} \sum_j p_{j,k} &\quad \forall i,k \in [1,n] \\
    \sum_k \tfrac{1}{k} \sum_i p_{i,k} \leq 1 &
    \end{matrix}
    \right\}. \label{eq:delta}
\end{align}
All of the terms in the objective: $\pvec_k^1$, $\qvec_l^1$, $\pvec_k^0$, $\qvec_l^0$, $r_k$, $s_l$, $\Pcal(\zerovec)$, and $\Qcal(\zerovec)$ 
can be computed from $\Pbf$ and $\Qbf$.
\end{theorem}

Using the construction above, we reduce the number of optimized variables in the inner minimax from exponential size to just quadratic size. Note that the objective in Eq. \eqref{eq:dual-marginal} remains bilinear over the optimized variables ($\Pbf$ and $\Qbf$), as in the original formulation (Eq. \eqref{eq:dual}) that is bilinear over $\Pcal(\hat{\Yvec}) $ and $\Qcal(\check{\Yvec})$.

\subsection{Optimization}
\label{sec:optimization}

One of the benefits of optimizing a loss metric using the adversarial prediction framework is that the resulting dual optimization (e.g., Eq. \eqref{eq:dual} and Eq. \eqref{eq:dual-marginal}) is convex (or concave in our case of optimizing performance metric) in $\theta$, 
despite the original metric that we want to optimize is non-convex and non-continuous. Therefore, to achieve the global solution of $\theta$, we can just use the standard gradient ascent algorithm. The sub-gradient of the objective with respect to theta is described in the following theorem.
\begin{theorem}
\label{thm:gradient}
Let $\Lcal(\theta)$ be the objective of the maximization over $\theta$ in Eq. \eqref{eq:dual-marginal} and let $\Qbf^*$ be the solution of the inner minimization over $\Qbf$, then the sub-gradient of $-\Lcal(\theta)$ with respect to $\theta$ includes:
\begin{align}
    \partial_{\theta} -\Lcal(\theta) \ni \Psi \left( {\Qbf^*}^\intercal \onevec - \yvec\right). \label{eq:gradient}
\end{align}
\end{theorem}

To solve the inner minimax over $\Qvec$ and $\Pvec$, we eliminate the inner-most optimization over $\Pvec$ by introducing dual variables over some of the constraints on $\Pvec$ and a slack variable that convert maximization into a set of constraints over $\Qvec$ and the slack variable. This results in 
a linear program optimization problem.
\begin{theorem}
\label{thm:lp}
The inner minimization over $\Qbf$ in Eq. \eqref{eq:dual-marginal} can be solved as a linear program in the form of:
\begin{align}
    &\min_{\Qbf \in \Delta; \alphavec \ge 0; v \geq 0}
     \;
      v + c(\Qbf) - \langle \Qbf,  \Psi^\intercal \theta \onevec^\intercal \rangle \label{eq:lp} \\
    \text{s.t.: } & 
    v \geq \Zbf(\Qbf)_{(i,k)} - \alpha_{i,k} + \tfrac{1}{k} \textstyle\sum_{j} \alpha_{j,k}, \quad \forall i,k \in [1,n], \notag
\end{align}
where  $c(\Qbf)$ is a linear function of $\Qbf$ and $\Zbf(\Qbf)$ is a matrix-valued linear function of $\Qbf$, both of which are defined analytically by the form of the metric.\footnote{Please see Appendix \ref{sec:proof-appendix} for the details.}
\end{theorem}

\subsection{Metric Constraints}

In some machine learning settings, we may want to optimize a performance metric subject to constraints on other metrics. This occurs in the case where there are trade-offs between different performance metrics. For example, a machine learning system may want to optimize the precision of the prediction, subject to its recall is greater than some threshold. For these tasks, we write the adversarial prediction formulation as:
\begin{align}
    &\max_{\Pcal(\hat{\Yvec}) } \; \min_{\Qcal(\check{\Yvec}) 
    } \; 
    \mathbb{E}_{\tilde{P}(\Xbf); \Pcal(\hat{\Yvec});
    	\Qcal(\check{\Yvec})} \left[\text{metric}^{(0)}(\hat{\Yvec}, \check{\Yvec}) \right] \label{eq:primal-cs}\\ 
    &\text{s.t.:  } \mathbb{E}_{\tilde{P}(\Xbf); 
    	\Qcal(\check{\Yvec})}[\phi({\bf X},\check{\Yvec})]
    = \mathbb{E}_{\tilde{P}(\Xbf,\Ybf)}\left[\phi({\bf X},{\Yvec}) \right], \nonumber  \\
     &  \qquad \mathbb{E}_{\tilde{P}(\Xbf, \Ybf); \Pcal(\hat{\Yvec})}
    	\left[ \text{metric}^{(i)}(\hat{\Yvec}, \Yvec) \right] \geq \tau_i, \; \forall i \in [1,t],  \notag 
\end{align}
where $t$ is the number of metric constraints.
In this formulation, we constraint the predictor to choose a conditional distribution in which the expected values of the constraint metrics evaluated on empirical data are greater than some threshold $\tau$. 

As in Section \ref{sec:formulation}, we use a marginalization technique to reduce the size of the optimization problem as stated in Theorem \ref{thm:theta_q_p_cs}.
\begin{theorem}
\label{thm:theta_q_p_cs}
Let $\Pbf$ and $\Qbf$ be the marginal predictor and adversary probability matrices respectively. Given a performance metric that follows the construction in Eq. \eqref{eq:metric}, a set of constraints over metrics that also follows the construction in Eq. \eqref{eq:metric},
and features that are additive over each sample, the dual optimization formulation of (Eq. \eqref{eq:primal-cs}) can be computed as:
\begin{flalign}
	&\max_{\theta} \! \bigg\{ \!\!
     \min_{\Qbf \in \Delta}
     \max_{\Pbf \in \Delta \cap \Gamma} \! \bigg[\!
      \sum_{k,l \in [0,n]} \!\! \sum_j \tfrac{1}{g_j^{(0)}\!(k, l)} \!\Big\{ a_j^{(0)}  [ \pvec_k^1 \!\cdot\! \qvec_l^1]\! \label{eq:dual-marginal-cs} \\
     &\!+\! b_j^{(0)}  [ \pvec_k^0 \cdot \qvec_l^0]  \!+\! f_j^{(0)} \! (k,l) r_k s_l \Big\} \! - \! \langle \Qbf^\intercal \onevec,\! \Psi^\intercal \theta \rangle \bigg] \!+\!  \langle \yvec,\! \Psi^\intercal \theta \rangle \!\bigg\}\!, \notag
\end{flalign}
where $\Delta$ is the set of marginal probability matrices defined in Eq \eqref{eq:delta}, and $\Gamma$ is the set of marginal probability matrices defined as:
\begin{align}
    &\Gamma \!=\! \Bigg\{ \!\Pvec \Bigg\vert \sum_{k \in [0,n]} \sum_j \tfrac{1}{g_j^{(i)}(k, l)} \Big\{ a_j^{(i)} [\pvec_k^1 \!\cdot\! \yvec] + b_j^{(i)} [\pvec_k^0 \!\cdot\! (1 \!-\! \yvec)] \notag \\
    &  +\! f_j^{(0)} \! (k,l) r_k  \Big\} \geq \tau_i,  \forall i \!\in\! [1,t]
     \Bigg\} \!,  \text{\small where } l \!=\! \textstyle\sum_{i'} y_{i'}.
\end{align}
All of the terms in the objective: $\pvec_k^1$, $\qvec_l^1$, $\pvec_k^0$, $\qvec_l^0$, $r_k$, $s_l$, $\Pcal(\zerovec)$, and $\Qcal(\zerovec)$, 
can be computed from $\Pbf$ and $\Qbf$.
\end{theorem}

Note that the resulting optimization in the case where we have metric constraints (Eq. \eqref{eq:dual-marginal-cs}) is relatively similar with the standard case (Eq. \eqref{eq:dual-marginal}). The only difference is the additional constraints over $\Pbf$. Since the constraints in the set $\Gamma$ are also just linear constraints over $\Pbf$, we can also rewrite the inner minimization over $\Qbf$ in Eq. \eqref{eq:dual-marginal-cs} as a linear program.
\begin{theorem}
\label{thm:lp-cs}
The inner minimization over $\Qbf$ in Eq. \eqref{eq:dual-marginal-cs} can be solved as a linear program
in the form of:
\begin{align}
    &\min_{\Qbf \in \Delta; \alphavec \ge 0; \betavec \ge 0; v \geq 0}
      v + c(\Qbf) \!-\! \langle \Qbf,  \Psi^\intercal \theta \onevec^\intercal \rangle + \textstyle\sum_l \left( \mu_l \!-\! \tau_l \right) \notag \\
    & \text{s.t.: } 
    v \geq \Zbf(\Qbf)_{(i,k)} \!-\! \alpha_{i,k} \!+\! \tfrac{1}{k} \textstyle\sum_{j} \alpha_{j,k}  \!+\! \textstyle\sum_l \beta_l \; (\Bbf^{(l)})_{(i,k)}  \notag\\ 
    &\forall i,k \in [1,n], \label{eq:lp-cs} 
\end{align}
where  $c(\Qbf)$ is a linear function of $\Qbf$ and $\Zbf(\Qbf)$ is a matrix-valued linear function of $\Qbf$, both of which are defined analytically by the form of the metric; whereas $\mu_l$ is a constant and $\Bbf^{(l)}$ is a matrix, both of which are defined analytically by the $l$-th metric constraint and the ground truth label.
\end{theorem}

\subsection{Integration into Differentiable Learning}

In this section, we aim to integrate our formulation into differentiable learning pipelines with a focus on those based upon neural network architectures. First, we note that even though we have reduced the number of variables in our formulation from exponential to quadratic size, it is still too big for most neural network learning tasks since the number of examples is usually big. Therefore, when optimizing the inner minimax over $\Qbf$ and $\Pbf$, rather than optimizing over all samples, we perform optimization for every minibatch which limits the size of optimized variables into a relatively small quadratic size.
We introduce non-linearity into our model by using the last layer of neural networks model as the features that we use to constraints the adversary's distribution in Eq. \eqref{eq:primal}. Consequently, in the training process, we propagate back the gradient signal in Eq. \eqref{eq:gradient} to the network.

To enable  easy integration with machine learning pipelines, we develop a programming interface for writing a custom performance metric. This interface enables the user to write an arbitrary complex performance metric based on the entities in the confusion matrix. If the metric is valid according to our metric construction in Eq \eqref{eq:metric}, we create an expression tree that stores all the operations in the metric. This expression tree is then used when computing the objective and constraints in Eq. \eqref{eq:dual-marginal} and Eq. \eqref{eq:dual-marginal-cs} as well as the LP formulations in Eq. \eqref{eq:lp} and Eq. \eqref{eq:lp-cs}. 
We implement our method on top of Julia programming language \citep{bezanson2017julia} and its machine learning framework, FluxML \citep{innes2018fashionable}. However, our method can be implemented in any other languages and frameworks.
Figure \ref{fig:code2} provides a code example for writing the definition of Cohen's kappa score metric. Note that our programming interface can handle a relatively complex performance metric. Figure \ref{fig:code3} shows an example where we want to optimize precision, with a constraint that the recall has to be greater than 0.8. For more examples of the code for various performance metrics, we refer the reader to Appendix \ref{sec:code-appendix}.

\begin{figure}[tb]
\begin{minted}[frame=lines,
               framesep=2mm]{text}
@metric Kappa
function define(::Type{Kappa}, C::ConfusionMatrix)
    pe = (C.ap * C.pp + C.an * C.pn) / C.all^2
    num = (C.tp + C.tn) / C.all - pe
    den = 1 - pe
    return num / den
end  
kappa = Kappa()
special_case_positive!(kappa)
special_case_negative!(kappa)
\end{minted}
\vspace{-6pt}
\caption{Code example for Cohen's kappa score.}
\vspace{-6pt}
\label{fig:code2}
\end{figure}

\subsection{Linear Program Solver and Runtime}

As mentioned in Section \ref{sec:optimization}, the inner minimization in the dual optimization of the adversarial prediction framework can be reformulated as a linear program (LP), which can be solved using any off-the-shelf LP solver such as Gurobi, Mosek, and Clp. The number of variables and constraints in the LP is $O(m^2)$, where $m$ is the batch size. The worst-case complexity of solving a linear program is $O(n^3)$ using the interior point algorithm where $n$ the number of variables. Therefore, the worst-case complexity of solving for the LP is $O(m^6)$ (solvers that exploit sparsity may reduce the runtime).  

To reduce the runtime complexity of solving the resulting LP, we develop a customized solver using the alternating direction method of multipliers (ADMM) technique \citep{douglas1956numerical,glowinski1975approximation,boyd2011distributed}. 
This reduces the worst-case runtime complexity to just $O(m^3)$, where $m$ is the batch size. 
In practice, for a batch size of 25, our ADMM-based solver takes roughly 10 - 30 milliseconds to converge in a desktop PC with an Intel Core i7 processor.
While it is noticeably slower than the cross-entropy loss computation, it is still practical, since for reasonably sized networks, the  loss function computation is usually dominated by the computation of the previous layers.
We refer the reader to Appendix \ref{sec:admm-appendix} for the detailed formulation of our custom solver.

\begin{figure}[tb]
\begin{minted}[frame=lines,
               framesep=2mm]{text}
@metric PR
function define(::Type{PR}, C::ConfusionMatrix)
    return C.tp / C.pp
end   
function constraint(::Type{PR}, C::ConfusionMatrix)
    return C.tp / C.ap >= 0.8
end   
prec_rec = PR()
special_case_positive!(prec_rec)
cs_special_case_positive!(prec_rec, true)
\end{minted}
\vspace{-6pt}
\caption{Code example for precision metric with a constraint on recall metric.}
\vspace{-6pt}
\label{fig:code3}
\end{figure}

\subsection{Fisher Consistency Property}

The behavior of a learning algorithm in an ideal setting (i.e., where the algorithm is given access to the true population distribution, and it is allowed to be optimized over the set of all measurable hypothesis functions), provides a useful theoretical validation. Fisher consistency requires that the prediction model yields the Bayes optimal decision
boundary in this setting \citep{tewari2007consistency,liu2007fisher}
The Fisher consistency of the adversarial prediction framework has been established previously for decomposable metrics,  bipartite matching, and graphical model \citep{fathony2018consistent,fathony2018efficient,fathony2018distributionally}. We establish the consistency of our approach in the following theorem.

\begin{theorem}
\label{thm:consistency}
Given a performance metric that follows the construction in Eq. \eqref{eq:metric}, the adversarial prediction formulation in Eq. \eqref{eq:primal} is Fisher consistent if the algorithm is optimized over a set of functions that are additive over each sample and sum statistics, i.e., $h(\xvec,\yvec) = \sum_{i,k} \rho_{i,k}(\xvec_i, y_i, k) \Ibb[\sum_i y_i = k]$, provided that $\rho_{i,j}$ is allowed to be optimized over the set of all measurable functions on the individual input space $(\xvec_i, y_i)$.
\end{theorem}

\section{EXPERIMENTS}

To evaluate our approach, we apply our formulation to classification tasks on 20 different tabular datasets from UCI repository \citep{DuaUCI} and benchmark datasets \citep{chu2005gaussian}, as well as
image datasets from MNIST and Fashion MNIST. For the multiclass datasets, we transform them into binary classification tasks by selecting one or more classes as the positive label and the rest as the negative label.
We construct a multi-layer perceptron (MLP) with two hidden layers for the tabular datasets and a convolutional neural network for the image datasets.
We evaluate the prediction using 6 different metrics: accuracy, F1 score, F2 score, the geometric mean of precision and recall (GPR), Matthews correlation coefficient (MCC), and Cohen's kappa score. We also evaluate the prediction using metric constraints, specifically, we train our method to optimize precision given that the recall is greater than certain thresholds. We select two different thresholds for the recall, 0.8 and 0.6. We then measure the prediction using precision at recall equal to the given thresholds.

\begin{figure}[tb]
\centering
\includegraphics[width=0.45\textwidth]{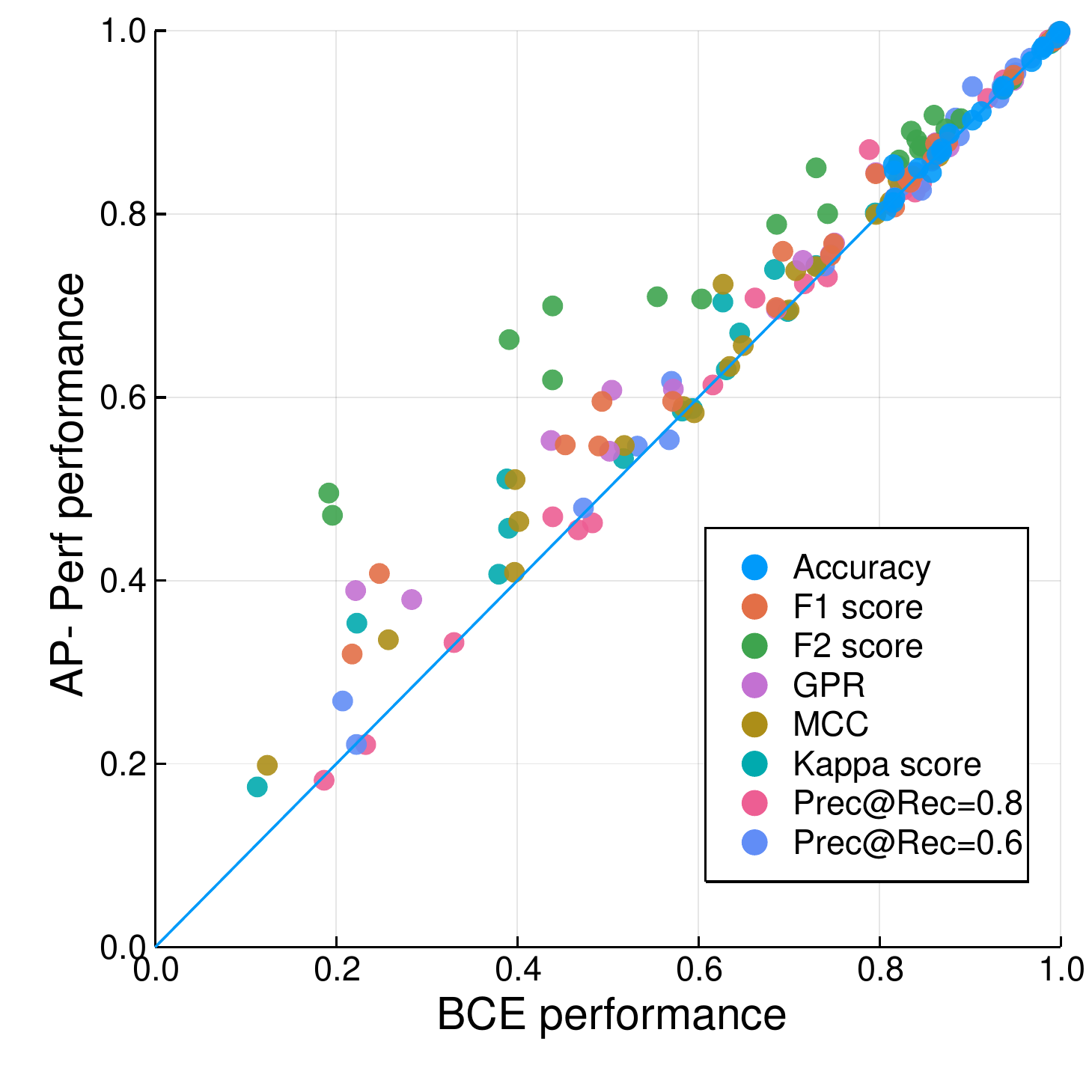}
\vspace{-8pt}
\caption{Comparison between BCE and AP-Perf. 
}
\vspace{-6pt}
\label{fig:scatter}
\end{figure}

We compare our method with the standard neural networks training that optimizes the binary cross-entropy (BCE) on the 22 datasets. In our experiment, we train our methods separately for each performance metric that we want to optimize, whereas for the BCE networks, we only train the networks once using the cross-entropy objective. 
We then measure the performance of the prediction using 8 metrics that we have selected.
For both methods, we perform a cross-validation using validation set to select the best L2 regularization among $\lambda = \{0, 0.001, 0.01, 0.1 \}$. 
In each dataset, we run the training procedure for 100 epochs.
After the training session finished, we compute the value of the metric for prediction in the testing dataset. 
For both methods, we select the predictive models that achieve the best metric in  the validation set.
We refer the reader to Appendix \ref{sec:experiment-appendix} for the details about the datasets and experiment setup.
The AP-Perf framework code is available at \href{https://github.com/rizalzaf/AdversarialPrediction.jl}{\small \texttt{https://github.com/ rizalzaf/\linebreak AdversarialPrediction.jl}}, whereas the experiment is available at   \href{https://github.com/rizalzaf/AP-examples}{\small \texttt{https://github.com/rizalzaf/AP-examples}}.

Figure \ref{fig:scatter} shows a scatter plot of the comparison between our method and the BCE on the 22 datasets. The x-axis in the plot denotes the BCE performance whereas the y-axis is the AP-Perf performance. The blue line in the plot denotes the case where the performance for both methods is equal. Points above the line indicate that AP-Perf outperforms the BCE on the particular dataset and evaluation metric. As we can see from the figure, almost all of the points in the scatter plot lie on or above the blue line. This shows the benefit of our methods in optimizing the performance metrics as opposed to  training the network using the cross-entropy objective. 
From the plot, we can infer that our method provides more benefit for the ``relatively hard problems'', i.e., the tasks where the BCE produces good accuracy but low to moderate performance in other metrics. We can also see that the AP-Perf consistently provides the best improvement over the BCE on the F2 score metric. This can be explained by the fact that the F2 score is the only imbalance metric from the list, i.e., it emphasizes some parts of the metric (in F2-score, recall is two times as important as precision). Since the BCE optimizes a proxy to a balanced metric (accuracy), it suffers more in the case where an imbalance metric is used for evaluation.

\section{CONCLUSION}

We developed a technique and programming interface that enable practitioners to integrate custom non-decomposable metric into differentiable learning. Our methods support a vast range of commonly used performance metrics. 
The list of metrics that our approach support is, however, far from exhaustive. The most noticeable missing metric is the area-based metric (e.g., AUC-ROC), which cannot be directly computed from the value of the entities in the confusion matrix, and ranking-based metrics (e.g., precision@k and MAP).
Our future works aim to close these gaps in the metric that we do not support.



{
\bibliographystyle{plainnat}
\bibliography{biblio}
}

\begin{appendices}
\input{supplementary.tex}

\end{appendices}

\end{document}

%% file: supplementary.tex
\newpage
\appendix
\onecolumn

\section{Proofs}
\label{sec:proof-appendix}

\subsection{Proof of Theorem \ref{thm:marginal}}

\begin{proof}[Proof of Theorem \ref{thm:marginal}]

The metric in Eq. \eqref{eq:metric} can be written in a variable notation as:
\begin{equation}
    \text{metric}(\yvechat,\yvec) = 
    \sum_j \frac{a_j \sum_i \yhat_i y_i + b_j \sum_i (1\!-\!\yhat_i) (1\!-\!y_i) + f_j(\sum_i \yhat_i, \sum_i y_i)}{g_j(\sum_i \yhat_i, \sum_i y_i)}.
\end{equation}
Therefore, the expected value of the metrics can be computed as:
\begin{align}
    &\mathbb{E}_{\Pcal(\hat{\Yvec});
    	\Qcal(\check{\Yvec})}\left[ \text{metric}(\hat{\Yvec},\check{\Yvec}) \right] \\
    \overset{a}{=}& \sum_{\yvechat \in \{0,1\}^n} \sum_{\yvecchk \in \{0,1\}^n} \Pcal(\yvechat) \Qcal(\yvecchk) \; \text{metric}(\yvechat,\yvecchk) \\
    \overset{b}{=}& \sum_{\yvechat \in \{0,1\}^n} \sum_{\yvecchk \in \{0,1\}^n} \Pcal(\yvechat) \Qcal(\yvecchk)   \sum_j \frac{a_j \sum_i \yhat_i \ychk_i + b_j \sum_i (1\!-\!\yhat_i) (1\!-\!\ychk_i) + f_j(\sum_i \yhat_i, \sum_i \ychk_i)}{g_j(\sum_i \yhat_i, \; \sum_i \ychk_i)} \\
    \overset{c}{=}& \sum_{k \in [0,n]} \sum_{l \in [0,n]} \sum_{ \left\{\yvechat \mid {\Sigma_i} \yhat_i = k \right\} } \sum_{ \left\{\yvecchk \mid {\Sigma_i} \ychk_i = l \right\} } \Pcal(\yvechat) \Qcal(\yvecchk) \bigg(  \sum_j \frac{a_j \sum_i \yhat_i \ychk_i + b_j \sum_i (1\!-\!\yhat_i) (1\!-\!\ychk_i) + f_j(k, l)}{g_j(k, \; l)} \bigg)\\
    \overset{d}{=}&  \sum_{k \in [0,n]} \sum_{l \in [0,n]} \sum_j \frac{1}{g_j(k, l)} \Big( a_j \textstyle \sum_{ \left\{\yvechat \mid {\Sigma_i} \yhat_i = k \right\} } \sum_{ \left\{\yvecchk \mid {\Sigma_i} \ychk_i = l \right\} } \Pcal(\yvechat) \Qcal(\yvecchk)  \sum_i \yhat_i \ychk_i  \\
    & \qquad \qquad  \;  + b_j \textstyle \sum_{ \left\{\yvechat \mid {\Sigma_i} \yhat_i = k \right\} } \sum_{ \left\{\yvecchk \mid {\Sigma_i} \ychk_i = l \right\} } \Pcal(\yvechat) \Qcal(\yvecchk) \sum_i (1\!-\!\yhat_i) (1\!-\!\ychk_i) \notag \\ 
    & \qquad \qquad  \;  + \textstyle \sum_{ \left\{\yvechat \mid {\Sigma_i} \yhat_i = k \right\} } \sum_{ \left\{\yvecchk \mid {\Sigma_i} \ychk_i = l \right\} } \Pcal(\yvechat) \Qcal(\yvecchk)  f_j (k, l)
    \Big) \notag \\
    \overset{e}{=}& \sum_{k \in [0,n]} \sum_{l \in [0,n]} \sum_j \frac{1}{g_j(k, l)} \Big( a_j \textstyle\sum_i \Pcal(\yhat_i=1, {\textstyle\sum_{i'}} \yhat_{i'} = k)  \Qcal(\ychk_i=1, {\textstyle\sum_{i'}} \ychk_{i'} = l) \label{eq:marginal-0-to-n}  \\
    & \qquad \qquad + b_j 
    \textstyle\sum_i (\Pcal(\yhat_i=0, {\textstyle\sum_{i'}} \yhat_{i'} = k)) ( \Qcal(\ychk_i=0, {\textstyle\sum_{i'}} \ychk_{i'} = l)) + f_j(k,l) \Pcal ({\textstyle\sum_i} \yhat_i = k)  \Qcal({\textstyle\sum_i} \ychk_i = l)  \Big)  \notag \\
    \overset{f}{=}& \sum_{k \in [0,n]} \sum_{l \in [0,n]} \sum_j \frac{1}{g_j(k, l)}  \big( a_j [ \pvec_k^1 \cdot \qvec_l^1] + b_j [ \pvec_k^0 \cdot \qvec_l^0] + f_j(k,l) r_k s_l  \big) .
\end{align}

The transformations above are explained as follow:
\begin{enumerate}[(a)]
    \item Expanding the definition of expectation of the metric to the sum of probability-weighted metrics.
    \item Applying the construction of our performance metric.
    \item Grouping the values of the metric in terms of $\sum_i \yhat_i = k$ and $\sum_i \ychk_i = l$. 
    \item  Since each $f_j$ is just a linear function over $\sum_i \yhat_i \ychk_i$ and $\sum_i (1\!-\!\yhat_i) (1\!-\!\ychk_i)$, we can push the summation over $ \sum_{ \left\{\yvechat \mid {\Sigma_i} \yhat_i = k \right\} } \sum_{ \left\{\yvecchk \mid {\Sigma_i} \ychk_i = l \right\} }$ inside $f_j$.
    \item Since $\sum_i \yhat_i \ychk_i$ and $\sum_i (1\!-\!\yhat_i) (1\!-\!\ychk_i)$ are both decomposable, then the expectation over $\Pcal(\yvechat)$ and $ \Qcal(\yvecchk)$ for the case where $\sum_i \yhat_i = k$ and $\sum_i \ychk_i = l$ can be decomposed into each individual marginal probabilities $\Pcal(\yhat_i, \sum_{i'} \yhat_{i'} = k)$ and $\Qcal(\ychk_i, \sum_{i'} \ychk_{i'} = l)$. Similarly, given fixed $k$ and $l$, $f_j(k,l)$ is just a constant. Hence we can simplify the expectation over $f_j(k,l)$ in terms of the marginal probabilities of $\Pcal(\sum_{i} \yhat_{i} = k)$ and $\Qcal( \sum_{i} \ychk_{i} = l)$.
    \item Rewriting the marginal probabilities in vector notations.
\end{enumerate}

\end{proof}

\subsection{Proof of Theorem \ref{thm:theta_q_p}}
\label{sec:dual-marginal-appendix}

\begin{proof}[Proof of Theorem \ref{thm:theta_q_p}]
From Theorem \ref{thm:marginal} we know that:
\begin{align}
	&\max_{\theta}
     \mathbb{E}_{\tilde{P}(\Xbf, \Ybf)} \bigg[
     \min_{\Qcal(\check{\Yvec})}
     \max_{\Pcal(\hat{\Yvec}) }
    \mathbb{E}_{\Pcal(\hat{\Yvec});
    	\Qcal(\check{\Yvec})}\Big[ \text{metric}(\hat{\Yvec},\check{\Yvec}) - \theta^\intercal \left( \phi({\bf X},\check{\Yvec}) - \phi({\bf X},{\Yvec}) \Big)
    \right]
    \bigg] \\
    =& \max_{\theta}
     \mathbb{E}_{\tilde{P}(\Xbf, \Ybf)} \bigg[
     \min_{\Qcal(\check{\Yvec})}
     \max_{\Pcal(\hat{\Yvec}) } 
     \bigg[ \sum_{k \in [0,n]} \sum_{l \in [0,n]} \sum_j \tfrac{1}{g_j(k, l)}  \left\{ a_j [ \pvec_k^1 \cdot \qvec_l^1] + b_j [ \pvec_k^0 \cdot \qvec_l^0]  + f_j(k,l) r_k s_l \right\} \\
     & \qquad \qquad \qquad- 
    \mathbb{E}_{\Qcal(\check{\Yvec})}\Big[ \theta^\intercal \left( \phi({\bf X},\check{\Yvec}) - \phi({\bf X},{\Yvec}) \Big)
    \right]
    \bigg]. \notag
\end{align}

Note that the values for some $\pvec_k^a$ and $\qvec_l^a$ are known, i.e.:
\begin{align}
    (\pvec_0^1)_i &= \Pcal(\yhat_i=1, {\textstyle\sum_i} \yhat_i = 0)) = 0, \quad \forall i \in [1,n] \\
    (\pvec_n^0)_i &= \Pcal(\yhat_i=0, {\textstyle\sum_i} \yhat_i = n)) = 0, \quad \forall i \in [1,n] \\
    (\pvec_n^1)_i &= \Pcal(\yhat_i=1, {\textstyle\sum_i} \yhat_i = n)) = \Pcal(\onevec), \quad \forall i \in [1,n] \\
    (\pvec_0^0)_i &= \Pcal(\yhat_i=0, {\textstyle\sum_i} \yhat_i = 0)) = \Pcal(\zerovec), \quad \forall i \in [1,n]
\end{align}
and similarly for $\qvec_l^a$.

We now analyze the relation between $\pvec_k^1$ and $\pvec_k^0$ (which also applies to $\qvec_k^1$ and $\qvec_k^0$).
Note that each $\Pcal(\yvechat)$ such that $\sum_i \yhat_i = k$ appears $k$ times in $\sum_i \Pcal(\yhat_i=1, {\sum_i} \yhat_i = k))$, which implies:
\begin{align}
\textstyle r_k = \Pcal({\sum_i} \yhat_i = k) = \frac{1}{k} \sum_i \Pcal(\yhat_i=1, {\sum_i} \yhat_i = k)).
\end{align} 
Therefore, we also have the relation:
\begin{align}
    \Pcal(\yhat_i=0, {\textstyle \sum_i} \yhat_i = k)
    =&\textstyle \Pcal({\sum_i} \yhat_i = k) - \Pcal(\yhat_i=1, {\sum_i} \yhat_i = k) \notag \\
    =& \textstyle\frac{1}{k} \sum_i \Pcal(\yhat_i=1, {\sum_i}  \yhat_i = k)) - \Pcal(\yhat_i=1, {\sum_i} \yhat_i = k), \notag
\end{align}
for all $k \in [1, n\!-\!1]$. In vector notation, we can write:
\begin{align}
    r_k = & \tfrac{1}{k} (\pvec_{k}^1 \cdot \onevec) \\
    \pvec_k^0 =& \tfrac{1}{k} (\pvec_{k}^1 \cdot \onevec) \onevec - \pvec_{k}^1, \qquad \forall k \in [1, n\!-\!1].
\end{align}
We know already that $\pvec_n^0 = \zerovec$. For computing $\pvec_0^0$, we know that $\Pcal(\yhat_i=0, {\textstyle \sum_i} \yhat_i = 0) =\Pcal({\textstyle \sum_i} \yhat_i = 0) = \Pcal(\zerovec)$ which can be computed as:
\begin{align}
    \Pcal(\zerovec) &= \textstyle 1 - \sum_{k \in [1,n]} \Pcal(\sum_i \yhat_i = k) \\
    &= \textstyle 1- \sum_{k \in [1,n]} \frac{1}{k} \sum_i \Pcal(\yhat_i=1, {\sum_i} \yhat_i = k)) \notag \\
    &= \textstyle 1- \sum_{k \in [1,n]} \frac{\pvec_k^1 \cdot \onevec}{k} \notag
\end{align}

Therefore, we can compute all values in $\pvec_k^0, \forall k \in [0,n]$, $r_k$, $\Pcal(\zerovec)$, and $\Pcal(\onevec)$ from $\pvec_k^1$, and thus we can perform optimization over $\pvec_k^1$ and $\qvec_k^1$ only. For short, we write the as just $\pvec_k$ and $\qvec_k$. Note that we know that $\pvec_0 = \qvec_0 = \zerovec$. Therefore, it suffices to optimize only over $\pvec_k$ and $\qvec_k$, for all $k \in [1,n]$. Let us denote a $n \times n$ matrix $\Pbf$ where each column $\Pbf_{(:,k)}$ represents $\pvec_k$. Similarly, we denote a matrix $\Qbf$ for $\qvec_k$.

Let us take a look at the property of the marginal probability matrices $\Pbf$ and $\Qbf$. To be a valid marginal probability, $\Pbf$ has to satisfy the following constraints:
\begin{align}
    p_{i,k} \geq 0 &\quad \forall i,k \in [1,n] \\
    \textstyle\sum_{k} p_{i,k} \leq 1 &\quad \forall i \in [1,n] \\
     p_{i,k} \leq \tfrac{1}{k} \textstyle\sum_j p_{j,k} &\quad \forall i,k \in [1,n] \\
    \textstyle\sum_k \tfrac{1}{k} \textstyle\sum_i p_{i,k} \leq 1 &
\end{align}
The constraints above are described below:
\begin{itemize}[itemsep=2pt,topsep=2pt,partopsep=0pt]
    \item The first constraint is for the non-negativity of probability.
    \item The second constraint is from  $\Pcal(\yhat_i = 1) = \sum_k \Pcal(\yhat_i = 1, \sum_i \yhat_i = k) \leq 1$. 
    \item The third constraint comes from the fact that each $\Pcal(\yvechat)$ such that $\sum_i \yhat_i = k$ appears $k$ times in $\sum_i \Pcal(\yhat_i=1, {\sum_i} \yhat_i = k))$, and thus, $\Pcal({\sum_i} \yhat_i = k) = \frac{1}{k} \sum_i \Pcal(\yhat_i=1, {\sum_i} \yhat_i = k))$. Therefore, the inequality of $\Pcal(y_i = 1, {\sum_i} \yhat_i = k) \leq \Pcal({\sum_i} \yhat_i = k)$ must hold which implies the third constraint.
    \item The fourth constraint comes from the fact that $\sum_k \Pcal({\sum_i} \yhat_i = k) \leq 1$.
\end{itemize}
The same constraints also need to hold for the probability matrix $\Qbf$.
We can also see that satisfying the third and fourth constraints implies the second constraints, i.e.:
\begin{equation}
    \sum_k p_{i,k} \leq \sum_k \tfrac{1}{k} \textstyle\sum_j p_{j,k} \leq 1.
\end{equation}

Now we take a look at the features.
Let the pair $(\xvec, \yvec)$ be the empirical training data. Based on the construction of our features, we compute the potentials for $\theta^\intercal \phi(\xvec, \yvec)$ as:
\begin{equation}
    \theta^\intercal \phi(\xvec, \yvec) =  \theta^\intercal  \sum_i \phi(\xvec, y_i) = \theta^\intercal  \sum_i \Ibb[y_i =1] \phi(\xvec, y_i=1) = \langle \yvec, \Psi^\intercal \theta \rangle, 
\end{equation}
and the potentials for $\mathbb{E}_{\Qcal(\check{\Yvec})} \left[ \theta^\intercal \phi(\xvec, \check{\Yvec}) \right]$ as:
\begin{equation}
    \mathbb{E}_{\Qcal(\check{\Yvec})} \left[ \theta^\intercal \phi(\xvec, \check{\Yvec}) \right] = \mathbb{E}_{\Qcal(\check{\Yvec})} \left[ \theta^\intercal  \sum_i \phi(\xvec, \Ychk_i) \right]
    =  \theta^\intercal \sum_i \Qcal(\ychk_i = 1) \phi(\xvec, \ychk_i =1) = \langle \Qbf^\intercal \onevec, \Psi^\intercal \theta \rangle. 
\end{equation}

Therefore, we can simplify Eq. \eqref{eq:dual} as:
\begin{align}
	\max_{\theta} \Bigg\{ 
     \min_{\Qbf \in \Delta}
     \max_{\Pbf \in \Delta}  \Bigg[
      \sum_{k \in [0,n]} \sum_{l \in [0,n]} \sum_j \tfrac{1}{g_j(k, l)}  \left\{ a_j [ \pvec_k^1 \cdot \qvec_l^1] + b_j [ \pvec_k^0 \cdot \qvec_l^0]  + f_j(k,l) r_k s_l \right\} - \langle \Qbf^\intercal \onevec, \Psi^\intercal \theta \rangle \Bigg] +  \langle \yvec, \Psi^\intercal \theta \rangle \!\Bigg\}\!, 
\end{align}
where $\Delta$ is the set of valid marginal probability matrix denoted as:
\begin{align}
    \Delta = \left\{\Pvec \middle\vert 
    \begin{matrix}
    p_{i,k} \geq 0 &\quad \forall i,k \in [1,n] \\
    p_{i,k} \leq \tfrac{1}{k} \sum_j p_{j,k} &\quad \forall i,k \in [1,n] \\
    \sum_k \tfrac{1}{k} \sum_i p_{i,k} \leq 1 &
    \end{matrix}
    \right\}.
\end{align}

\end{proof}

\subsection{Proof of Theorem \ref{thm:gradient}}

\begin{proof}[Proof of Theorem \ref{thm:gradient}]
The result follows directly from the rule of subgradient of maximum function.
\begin{align}
	-\Lcal(\theta) &= 
     \max_{\Qbf \in \Delta}
     \min_{\Pbf \in \Delta}  \Bigg[
      -\sum_{k \in [0,n]} \sum_{l \in [0,n]} \sum_j \tfrac{1}{g_j(k, l)}  \left\{ a_j [ \pvec_k^1 \cdot \qvec_l^1] + b_j [ \pvec_k^0 \cdot \qvec_l^0]  + f_j(k,l) r_k s_l \right\}  + \langle \Qbf^\intercal \onevec, \Psi^\intercal \theta \rangle \Bigg] - \langle \yvec, \Psi^\intercal \theta \rangle 
\end{align}
\begin{align}
    \partial_{\theta} -\Lcal(\theta) &\ni \Psi \left( {\Qbf^*}^\intercal \onevec - \yvec\right), \; \text{where:} \\
   {\Qbf^*} &= \argmax_{\Qbf \in \Delta}
     \min_{\Pbf \in \Delta}  \Bigg[
      -\sum_{k \in [0,n]} \sum_{l \in [0,n]} \sum_j \tfrac{1}{g_j(k, l)}  \left\{ a_j [ \pvec_k^1 \cdot \qvec_l^1] + b_j [ \pvec_k^0 \cdot \qvec_l^0]  + f_j(k,l) r_k s_l \right\}  + \langle \Qbf^\intercal \onevec, \Psi^\intercal \theta \rangle \Bigg]  \notag
\end{align}
\end{proof}

\subsection{Proof of Theorem \ref{thm:lp}}

\label{sec:thm-lp-appendix}

\begin{proof}[Proof of Theorem \ref{thm:lp}]
The inner minimization over $\Qbf$ in Eq. \eqref{eq:dual-marginal} is:
\begin{align}
    \min_{\Qbf \in \Delta}
     \max_{\Pbf \in \Delta}  \Bigg[
      \sum_{k \in [0,n]} \sum_{l \in [0,n]} \sum_j \tfrac{1}{g_j(k, l)}  \left\{ a_j [ \pvec_k^1 \cdot \qvec_l^1] + b_j [ \pvec_k^0 \cdot \qvec_l^0]  + f_j(k,l) r_k s_l \right\}  - \langle \Qbf^\intercal \onevec, \Psi^\intercal \theta \rangle \Bigg]. \label{eq:inner-q}
\end{align}

Denote:
\begin{equation}
    \Ocal(\Qbf, \Pbf) = \sum_{k \in [0,n]} \sum_{l \in [0,n]} \sum_j \tfrac{1}{g_j(k, l)}  \left\{ a_j [ \pvec_k^1 \cdot \qvec_l^1] + b_j [ \pvec_k^0 \cdot \qvec_l^0]  + f_j(k,l) r_k s_l \right\}.
\end{equation}
Since the objective in $\Ocal(\Qbf, \Pbf)$ is a bilinear function over $\Qbf$ and $\Pbf$, it can be written in the form of $\Ocal(\Qbf, \Pbf) = \left\langle  \frac{\partial \Ocal(\Qbf, \Pbf)}{\partial \Pbf}, \Pbf \right\rangle + c(\Qbf)$, where $c(\Qbf)$ is the terms that are constant over $\Pbf$.
Therefore,
Eq. \eqref{eq:inner-q} can be written as:
\begin{align}
    \min_{\Qbf \in \Delta}
     \max_{\Pbf \in \Delta} \;
      \left\langle \Zbf(\Qbf), \Pbf \right\rangle  + c(\Qbf)  - \langle \Qbf, \Wbf \rangle,
\end{align}
where $\Zbf(\Qbf) = \frac{\partial \Ocal(\Qbf, \Pbf)}{\partial \Pbf}$, and $\Wbf = \Psi^\intercal \theta \onevec^\intercal $. Note that both $\Zbf(\Qbf)$ and $c(\Qbf)$ are some linear functions that depend on the metric.

We expand the constraints over $\Pbf$ as:
\begin{align}
    &\min_{\Qbf \in \Delta}
     \max_{\Pbf} \;
      \left\langle \Zbf(\Qbf), \Pbf \right\rangle + c(\Qbf) - \langle \Qbf, \Wbf \rangle \\
    \text{s.t.: } & 
    p_{i,k} \geq 0 \quad \forall i,k \in [1,n] \notag \\
    & p_{i,k} \leq \tfrac{1}{k} \textstyle\sum_j p_{j,k} \quad \forall i,k \in [1,n] \notag \\
    & \textstyle\sum_k \tfrac{1}{k} \sum_i p_{i,k} \leq 1  \notag
\end{align}

We now perform a change of variable.
Let us transform $\Pbf$ to a matrix $\Abf$ where its element contains the value of $a_{i,k} = \frac{1}{k} a_{i,k}$. We can rewrite the objective as:
\begin{align}
    &\min_{\Qbf \in \Delta}
     \max_{\Abf} \;
      \left\langle \Zbf'(\Qbf), \Abf \right\rangle  + c(\Qbf)  - \langle \Qbf, \Wbf \rangle \\
    \text{s.t.: } & 
    a_{i,k} \geq 0 \quad \forall i,k \in [1,n] \notag \\
    & a_{i,k} \leq \tfrac{1}{k} \textstyle\sum_j a_{j,k} \quad \forall i,k \in [1,n] \notag \\
    & \textstyle\sum_k \sum_i a_{i,k} \leq 1  \notag,
\end{align}
where $\Zbf'(\Qbf)$ is the linearly transformed $\Zbf(\Qbf)$ to adjust the transformation of the variable from $\Pbf$ to $\Abf$.

Using duality, we introduce a Lagrange dual variable for $a_{i,k} \leq \tfrac{1}{k} \textstyle\sum_j a_{j,k} $  constraint.
\begin{align}
    &\min_{\Qbf \in \Delta; \alphavec \ge 0}
     \max_{\Abf} \;
      \left\langle \Zbf'(\Qbf), \Abf \right\rangle + c(\Qbf) - \langle \Qbf, \Wbf \rangle - \sum_{i,k} \alpha_{i,k} \left( a_{ik} - \tfrac{1}{k}  \textstyle\sum_j a_{j,k} \right) \\
    \text{s.t.: } & 
    a_{i,k} \geq 0 \quad \forall i,k \in [1,n] \notag \\
    & \textstyle\sum_k \sum_i a_{i,k} \leq 1  \notag
\end{align}

We regroup the terms that depend on $\Abf$ as:
\begin{align}
    &\min_{\Qbf \in \Delta; \alphavec \ge 0}
     \max_{\Abf} \;
      \left\langle \Zbf'(\Qbf), \Abf \right\rangle 
      - \sum_{i,k} a_{i,k} \left( \alpha_{ik} - \tfrac{1}{k}  \textstyle\sum_j \alpha_{j,k} \right)
      + c(\Qbf) - \langle \Qbf, \Wbf \rangle  \\
    \text{s.t.: } & 
    a_{i,k} \geq 0 \quad \forall i,k \in [1,n] \notag \\
    & \textstyle\sum_k \sum_i a_{i,k} \leq 1  \notag
\end{align}

We now eliminate the inner maximization over $\Abf$ by transforming it into constraints as follows:
\begin{align}
    &\min_{\Qbf \in \Delta; \alphavec \ge 0; v}
     \;
      v + c(\Qbf) - \langle \Qbf, \Wbf \rangle \\
    \text{s.t.: } & 
    v \geq 0 \notag \\
    &v \geq (\Zbf'(\Qbf))_{(i,k)} - \alpha_{i,k} + \tfrac{1}{k} \textstyle\sum_{j} \alpha_{j,k}, \qquad \forall i,k \in [1,n]. \notag
\end{align}

The formulation above can be written in a standard linear program as:
\begin{align}
    &\min_{\Qbf; \alphavec; v}
     \;
      v + c(\Qbf) - \langle \Qbf, \Wbf \rangle  \\
    \text{s.t.: } & 
    q_{i,k} \geq 0 \quad \forall i,k \in [1,n] \notag \\
    &\alpha_{i,k} \geq 0 \quad \forall i,k \in [1,n] \notag \\
    &v \geq 0 \notag \\
    & q_{i,k} \leq \tfrac{1}{k} \textstyle\sum_j q_{j,k} \quad \forall i,k \in [1,n] \notag \\
    & \textstyle\sum_k \tfrac{1}{k} \sum_i q_{i,k} \leq 1  \notag \\
    &v \geq (\Zbf'(\Qbf))_{(i,k)} - \alpha_{i,k} + \tfrac{1}{k} \textstyle\sum_{j} \alpha_{j,k}, \qquad \forall i,k \in [1,n], \notag
\end{align}
where  $c(\Qbf)$ is a linear function of $\Qbf$ and $\Zbf'(\Qbf)$ is a matrix-valued linear function of $\Qbf$, both of which are defined analytically by the form of the metric.
\end{proof}

\subsection{Proof of Theorem \ref{thm:theta_q_p_cs}}

\begin{proof}[Proof of Theorem \ref{thm:theta_q_p_cs}]

Let us take a look at the expectation in the constraints:
\begin{align}
    &\mathbb{E}_{\Pcal(\hat{\Yvec})}\left[ \text{metric}(\hat{\Yvec},\Yvec) \right] \\
    =& \sum_{\yvechat \in \{0,1\}^n} \Pcal(\yvechat)  \; \text{metric}(\yvechat,\yvec) \\
    =& \sum_{\yvechat \in \{0,1\}^n} \Pcal(\yvechat) \sum_j \frac{a_j \sum_i \yhat_i y_i + b_j \sum_i (1\!-\!\yhat_i) (1\!-\!y_i) + f_j(\sum_i \yhat_i, \sum_i y_i)}{g_j(\sum_i \yhat_i, \; \sum_i y_i)}  \\
    =&\sum_{k \in [0,n]} \sum_j \frac{a_j \textstyle \sum_{ \left\{\yvechat \mid {\Sigma_i} \yhat_i = k \right\} }  \Pcal(\yvechat)  \sum_i \yhat_i y_i + b_j \textstyle \sum_{ \left\{\yvechat \mid {\Sigma_i} \yhat_i = k \right\} } \Pcal(\yvechat) \sum_i (1\!-\!\yhat_i) (1\!-\!y_i) + \textstyle \sum_{ \left\{\yvechat \mid {\Sigma_i} \yhat_i = k \right\} } \Pcal(\yvechat) f_j(k, l ) }{g_j(k, l)}  \\
    =&\sum_{k \in [0,n]} \sum_j \frac{a_j \textstyle \sum_i \Pcal(\yhat_i=1, {\textstyle\sum_{i'}} \yhat_{i'} = k) y_i + b_j \textstyle \sum_i \Pcal(\yhat_i=0, {\textstyle\sum_{i'}} \yhat_{i'} = k) (1\!-\!y_i) + \textstyle \sum_i \Pcal({\textstyle\sum_i} \yhat_i = k) f_j(k, l )}{g_j(k, l)}  \\
    =&\sum_{k \in [0,n]} \sum_j \frac{ a_j [\pvec_k^1 \cdot \yvec] + b_j [\pvec_k^0 \cdot (1 \!-\! \yvec)] +  f_j(k,l) r_k}{g_j(k, l)} 
\end{align}
where $l = \sum_i y_i$. 
Therefore, 
the metric constraints can be written as:
\begin{align}
    &\sum_{k \in [0,n]} \sum_j \frac{  a_j [\pvec_k^1 \cdot \yvec] + b_j [\pvec_k^0 \cdot (1 \!-\! \yvec)] +  f_j(k,l) r_k}{g_j(k, l)}  \geq \tau_i, \; \forall i \in [1,t] \notag
\end{align}

The dual formulation of Eq. \eqref{eq:primal-cs} is:
\begin{align}
	\max_{\theta} \; &
     \mathbb{E}_{\tilde{P}(\Xbf,\Ybf)} \left[
     \min_{\Qcal(\check{\Yvec})}
     \max_{\Pcal(\hat{\Yvec})  \in \Gamma}
    \mathbb{E}_{\Pcal(\hat{\Yvec});\Qcal(\check{\Yvec})}\left[ \text{metric}^{(0)}(\hat{\Yvec},\check{\Yvec})
    + \theta^\intercal \left( \phi({\bf X},\check{\Yvec}) - \phi({\bf X},{\Yvec}) \right)
    \right]
    \right] \label{eq:dual-cs} \notag \\
    &\text{where : } \Gamma \triangleq \left\{ \Pcal(\hat{\Yvec})  \mid \mathbb{E}_{\tilde{P}(\Xbf,\Ybf);
    	\Pcal(\hat{\Yvec})}
    	\left[ \text{metric}^{(i)}(\hat{\Yvec}, \Yvec) \right] \geq \tau_i, \; \forall i \in [1,t] \right\} .
\end{align}

Following the analysis in the proof of Theorem \ref{thm:theta_q_p}, the dual formulation can be simplified as:
\begin{align}
	&\max_{\theta} \Bigg\{ 
     \min_{\Qbf \in \Delta}
     \max_{\Pbf \in \Delta \cap \Gamma}  \Bigg[
      \sum_{k \in [0,n]} \sum_{l \in [0,n]} \sum_j \tfrac{1}{g_j^{(0)}(k, l)} \Big\{  a_j^{(0)} [ \pvec_k^1 \cdot \qvec_l^1] + b_j^{(0)} [\pvec_k^0 \cdot \qvec_l^0] + f_j^{(0)} (k, l) r_k s_l \Big\}  - \langle \Qbf^\intercal \onevec, \Psi^\intercal \theta \rangle \Bigg] +  \langle \yvec, \Psi^\intercal \theta \rangle \!\Bigg\}\!, \notag 
\end{align}
where:
\begin{align}
    \Delta = \left\{\Pvec \middle\vert 
    \begin{matrix}
    p_{i,k} \geq 0 &\quad \forall i,k \in [1,n] \\
    p_{i,k} \leq \tfrac{1}{k} \sum_j p_{j,k} &\quad \forall i,k \in [1,n] \\
    \sum_k \tfrac{1}{k} \sum_i p_{i,k} \leq 1 &
    \end{matrix}
    \right\}, \; \text{and}
\end{align}
\begin{align}
    \Gamma =\Bigg\{ \Pvec \Bigg\vert& \sum_{k \in [0,n]} \sum_j \frac{ a_j^{(i)} [\pvec_k^1 \cdot \yvec] + b_j^{(i)} [\pvec_k^0 \cdot (1 \!-\! \yvec)] + f_j^{(i)} (k, l) r_k )}{g_j^{(i)}(k, l)} \geq \tau_i, \; \forall i \in [1,t]
     \Bigg\}, \; \text{where } l = \textstyle\sum_{i'} y_{i'}.
\end{align}

\end{proof}

\subsection{Proof of Theorem \ref{thm:lp-cs}}

\begin{proof}[Proof of Theorem \ref{thm:lp-cs}]
The inner minimization over $\Qbf$ in Eq. \eqref{eq:dual-marginal-cs} is relatively similar to the standard case (Eq. \eqref{eq:dual-marginal}). The only difference is the additional constraints over $\Pbf$.
Since the numerators of the metrics in the constraints are linear in terms of $\pvec_k^1$ and $\pvec_k^0$ (which also means linear in terms of $\pvec_k$), then the constraints in $\Gamma$ can be represented by some matrix $\Bbf^{(i)}$ and some constant $\mu_i$ such that:
\begin{align}
    \langle \Bbf^{(i)}, \Pbf \rangle + \mu_i  \geq \tau_i, \qquad \text{or, } \qquad \textstyle\sum_k (\bvec_k^{(i)})^\intercal \pvec_k^{(i)}  + \mu_i \geq \tau_i, \qquad \forall i \in  [1,t]
\end{align}
Following the change of variable in the proof of Theorem \ref{thm:lp}, we can also represent the constraint in terms of $\Abf$ using some matrix $\Bbf'^{(i)}$ such that:
\begin{align}
    \langle \Bbf'^{(i)}, \Abf \rangle + \mu_i  \geq \tau_i, \qquad \text{or, } \qquad \textstyle\sum_k (\bvec'^{(i)}_k)^\intercal \avec_k^{(i)}  + \mu_i \geq \tau_i, \qquad \forall i \in  [1,t]
\end{align}
Therefore, we have an inner optimization over $\Qbf$ and $\Abf$, which can be written as:
\begin{align}
    &\min_{\Qbf \in \Delta}
     \max_{\Abf} \;
      \left\langle \Zbf'(\Qbf), \Abf \right\rangle  + c(\Qbf)  - \langle \Qbf, \Wbf \rangle \\
    \text{s.t.: } & 
    a_{i,k} \geq 0 \quad \forall i,k \in [1,n] \notag \\
    & a_{i,k} \leq \tfrac{1}{k} \textstyle\sum_j a_{j,k} \quad \forall i,k \in [1,n] \notag \\
    & \textstyle\sum_k \sum_i a_{i,k} \leq 1  \notag \\
    & \langle \Bbf'^{(l)}, \Abf \rangle + \mu_l  \geq \tau_l, \forall l \in  [1,t] \notag
\end{align}

Using duality, we introduce Lagrange dual variables.
\begin{align}
    &\min_{\Qbf \in \Delta; \alphavec \ge 0; \betavec \ge 0}
     \max_{\Abf} \;
      \left\langle \Zbf'(\Qbf), \Abf \right\rangle + c - \langle \Qbf, \Wbf \rangle - \sum_{i,k} \alpha_{i,k} \left( a_{ik} - \tfrac{1}{k}  \textstyle\sum_j a_{j,k} \right) + \sum_l \beta_l \left(  \langle \Bbf'^{(l)}, \Abf \rangle + \mu_l  - \tau_l \right)\\
    &\text{s.t.: } 
    a_{i,k} \geq 0 \quad \forall i,k \in [1,n] \notag \\
    & \qquad \textstyle\sum_k \sum_i a_{i,k} \leq 1  \notag
\end{align}

We can convert the optimization in a standard linear program format as follows: 
\begin{align}
    &\min_{\Qbf; \alphavec; \betavec; v}
     \;
      v + c(\Qbf) - \langle \Qbf, \Wbf \rangle + \sum_l \left( \mu_l - \tau_l \right)  \\
    \text{s.t.: } & 
    q_{i,k} \geq 0 \quad \forall i,k \in [1,n] \notag \\
    &\alpha_{i,k} \geq 0 \quad \forall i,k \in [1,n] \notag \\
    &\beta_{l} \geq 0 \quad \forall l \in [1,s] \notag \\
    &v \geq 0 \notag \\
    & q_{i,k} \leq \tfrac{1}{k} \textstyle\sum_j q_{j,k} \quad \forall i,k \in [1,n] \notag \\
    & \textstyle\sum_k \tfrac{1}{k} \sum_i q_{i,k} \leq 1  \notag \\
    &v \geq (\Zbf'(\Qbf))_{(i,k)} - \alpha_{i,k} + \tfrac{1}{k} \textstyle\sum_{j} \alpha_{j,k} + \sum_l \beta_l \; (\Bbf'^{(l)})_{(i,k)}, \qquad \forall i,k \in [1,n] . \notag
\end{align}

\end{proof}

\subsection{Proof of Theorem \ref{thm:consistency}}

\begin{proof}[Proof of Theorem \ref{thm:consistency}]

Despite its apparent differences from
 standard empirical risk minimization (ERM), the dual formulation of the adversarial prediction (Eq. \eqref{eq:dual}) can be equivalently recast as an ERM method:

\begin{align}
	&\min_{\theta}
     \mathbb{E}_{\tilde{P}(\Xbf, \Ybf)} \left[ AL_{h_\theta}(\Xbf, \Ybf) \right], \qquad \text{where:} \\
      AL_{h_\theta}(\Xbf, \Ybf) & =
     \max_{\Qcal(\check{\Yvec})}
     \min_{\Pcal(\hat{\Yvec}) }
    \mathbb{E}_{\Pcal(\hat{\Yvec});\Qcal(\check{\Yvec})}\Big[ -\text{metric}(\hat{\Yvec},\check{\Yvec}) + h_\theta({\bf X},\check{\Yvec}) - h_\theta({\bf X},{\Yvec}) 
    \Big]
\end{align}
and $h_\theta(\xvec, \yvec) = \theta^\intercal \phi(\xvec,\yvec)$ is the Lagrangian potential function.
$AL_{h_\theta}(\xvec, \yvec) $ is then the surrogate loss for input $\xvec$ and 
label $\yvec$.
The Fisher consistency condition for the adversarial prediction can then be written as:

\begin{align}
\label{eq:def_consistency}
& h^* \in \Hcal^* \triangleq \argmin_{f} \mathbb{E}_{P(\Ybf|\xvec)} 
\sbr{\text{AL}_{h} (\xvec, \Ybf)}  
\\ \Rightarrow \ 
&\argmax_\yvec h^*(\xvec,\yvec) \subseteq  \argmax_{\yvec'} \mathbb{E}_{P(\Ybf|\xvec)} [\text{metric}(\yvec', \Ybf)]. \notag \end{align}

It has been shown by \citet{fathony2018consistent,fathony2018efficient}, for a given natural requirement of performance metric, i.e., $\text{metric}(\yvec,\yvec) > \text{metric}(\yvec,\yvec')$ for all $\yvec' \neq \yvec$, the adversarial prediction is Fisher consistent provided that $h$ is optimized over all measurable functions over the input space of $(\xvec, \yvec)$. We quote the result below:

\begin{proposition}[Consistency result from \citet{fathony2018consistent,fathony2018efficient}]
Suppose we have a metric that satisfy the natural requirement: $\text{metric}(\yvec,\yvec) > \text{metric}(\yvec,\yvec')$ for all $\yvec' \neq \yvec$. Then the adversarial surrogate loss $AL_h$ is Fisher consistent if $h$ is optimized over all measurable functions over the input space of $(\xvec, \yvec)$.
\end{proposition}

The key to the result above is the observation that given a loss metric $\text{loss}(\yvec', \yvec)$, for the optimal potential function $h^*$,
$h^*(\xvec,\yvec) + \text{loss}(\yvec^\diamond, \yvec)$ is invariant to $\yvec$ 
where $y^\diamond = \argmax_{\yvec'} \mathbb{E}_{P(\Ybf|\xvec)} [\text{metric}(\yvec', \Ybf)]$. This property is referred to as the \emph{loss reflective} property of the $h$ minimizer. For a performance metric, the property can be equivalently written as $h^*(\xvec,\yvec) - \text{metric}(\yvec^\diamond, \yvec)$ is invariant to $\yvec$.

We now want to reduce the input space that $h$ needs to operate in order to achieve to Fisher consistency property. We consider the restricted set of $h$ defined as: $h(\xvec,\yvec) = \sum_{i,k} \rho_{i,k}(\xvec, y_i, k) \Ibb[\sum_i y_i = k]$, where each $\rho_{\{i,k\}}$ is optimized over the set of all measurable functions on the individual input space of $(\xvec, y_i)$. If the performance metric follows the construction in Eq. \eqref{eq:metric}, then we can achieve the loss reflective property under the restricted set of $h$ by setting:
\begin{align}
    \rho_{i,k}(\xvec, y_i, k) = \sum_j \frac{a_j \sum_i y^\diamond_i y_i + b_j  \sum_i (1\!-\!y^\diamond_i) (1\!-\!y_i) +f_j( \sum_i y^\diamond_i, k)}{g_j(\sum_i y^\diamond_i, k)}.
\end{align}
This will render the loss reflective property as $h^*(\xvec,\yvec) - \text{metric}(\yvec^\diamond, \yvec) = \zerovec$.

Therefore, we can conclude that our method is Fisher consistent for a performance metric that follows the construction in Eq. \eqref{eq:metric} if the algorithm is optimized over a set of functions that are additive over each sample and sum statistics.
\end{proof}

\section{Experiment Details}
\label{sec:experiment-appendix}

\begin{table}[htb]
\centering
\caption{Properties of the datasets used in the experiments}
\label{tbl:dataset}
\begin{tabular}{@{}lrrrcc@{}}
\toprule
Dataset           & \multicolumn{1}{c}{\# train set} & \multicolumn{1}{c}{\# validation set} & \multicolumn{1}{c}{\# test set} & original classes & positive classes \\ \midrule
abalone           & 2,338                            & 585                                   & 1,254                           & {[}1,10{]}     & {[}6,10{]}     \\
adult             & 25,324                           & 6,331                                 & 13,567                          & {[}0,1{]}      & {[}1{]}        \\
appliancesenergy  & 11,051                           & 2,763                                 & 5,921                           & {[}0,1{]}      & {[}1{]}        \\
bankdomains2      & 4,587                            & 1,147                                 & 2,458                           & {[}1,10{]}     & {[}7,10{]}     \\
bankmarketing     & 25,318                           & 6,329                                 & 13,564                          & {[}0,1{]}      & {[}1{]}        \\
californiahousing & 11,558                           & 2,889                                 & 6,193                           & {[}1,10{]}     & {[}7,10{]}     \\
censusdomains     & 12,758                           & 3,190                                 & 6,836                           & {[}1,10{]}     & {[}7,10{]}     \\
computeractivity2 & 4,587                            & 1,147                                 & 2,458                           & {[}1,10{]}     & {[}8,10{]}     \\
default           & 16,800                           & 4,200                                 & 9,000                           & {[}0,1{]}      & {[}1{]}        \\
dutch             & 33,835                           & 8,459                                 & 18,126                          & {[}0,1{]}      & {[}1{]}        \\
eegeye            & 8,389                            & 2,097                                 & 4,494                           & {[}0,1{]}      & {[}1{]}        \\
fashion-mnist        & 48,000                           & 12,000                                 & 10,000                          & {[}0,9{]}      & {[}0{]}        \\
htru2             & 10,022                           & 2,506                                 & 5,370                           & {[}0,1{]}      & {[}1{]}        \\
letter            & 11,200                           & 2,800                                 & 6,000                           & {[}1,26{]}     & {[}22,26{]}    \\
mnist        & 48,000                           & 12,000                                 & 10,000                          & {[}0,9{]}      & {[}0{]}        \\
onlinenews        & 22,200                           & 5,550                                 & 11,894                          & {[}0,1{]}      & {[}1{]}        \\
pageblocks        & 3,065                            & 766                                   & 1,642                           & {[}1,5{]}      & {[}4,5{]}      \\
redwine           & 895                              & 224                                   & 480                             & {[}1,10{]}     & {[}7,10{]}     \\
sat               & 3,548                            & 887                                   & 2,000                           & {[}1,7{]}      & {[}6,7{]}      \\
sensorless        & 32,765                           & 8,191                                 & 17,553                          & {[}1,11{]}     & {[}7,10{]}     \\
shuttle           & 34,800                           & 8,700                                 & 14,500                          & {[}1,7{]}      & {[}4,7{]}      \\
whitewine         & 2,743                            & 686                                   & 1,469                           & {[}1,10{]}     & {[}7,10{]}     \\ \bottomrule
\end{tabular}
\end{table}

To evaluate our approach, we apply our formulation to classification tasks on 20 different tabular datasets from the UCI repository \citep{DuaUCI} and benchmark datasets \citep{chu2005gaussian}, as well as
image datasets from MNIST and Fashion MNIST. Table \ref{tbl:dataset} shows the list of the datasets and their properties (the number of samples in the train, validation, and test sets). Some of the datasets are binary classification tasks, which we use directly in our experiments. For the multiclass datasets, we transform them into binary classification tasks by selecting one or more classes as the positive label and the rest as the negative label.  Table \ref{tbl:dataset} also shows the original class labels in the dataset and the classes that we select as the positive label in the transformed binary classification. The distribution of the positive and negative samples in the training set of the resulting binary classification tasks is described in Table \ref{tbl:dataset-class}.
For all of the datasets, we perform standardization, i.e., transform all the variables into zero mean and unit variance. 
For the datasets that have not been divided into training and testing set, we split the data with the rule of 70\% samples for the train set and 30\% for the test set. In addition, during the training, we also split the original training set into two different sets, 80\% of the set for training, and the rest 20\% of the set for validation. 

\begin{table}[htb]
\centering
\caption{The number of positive and negative samples in the train set for each dataset}
\label{tbl:dataset-class}
\begin{tabular}{@{}lrrrr@{}}
\toprule
Dataset           & \multicolumn{1}{c}{\# train set} & \multicolumn{1}{c}{\# positive} & \multicolumn{1}{c}{\# negative} & \multicolumn{1}{c}{positive percentage} \\ \midrule
abalone           & 2338                      & 146                                     & 2192                                    & 6\% \hspace{1cm}                                        \\
adult             & 25324                     & 6258                                    & 19066                                   & 25\%  \hspace{1cm}                                      \\
appliancesenergy  & 11051                     & 2961                                    & 8090                                    & 27\%  \hspace{1cm}                                      \\
bankdomains2      & 4587                      & 1829                                    & 2758                                    & 40\%  \hspace{1cm}                                      \\
bankmarketing     & 25318                     & 2941                                    & 22377                                   & 12\%  \hspace{1cm}                                      \\
californiahousing & 11558                     & 4637                                    & 6921                                    & 40\%  \hspace{1cm}                                      \\
censusdomains     & 12758                     & 5088                                    & 7670                                    & 40\%  \hspace{1cm}                                      \\
computeractivity2 & 4587                      & 1379                                    & 3208                                    & 30\%  \hspace{1cm}                                      \\
default           & 16800                     & 3701                                    & 13099                                   & 22\% \hspace{1cm}                                       \\
dutch             & 33835                     & 17803                                   & 16032                                   & 53\% \hspace{1cm}                                       \\
eegeye            & 8389                      & 3769                                    & 4620                                    & 45\% \hspace{1cm}                                       \\
fashion-mnist     & 48000                     & 4764                                    & 43236                                   & 10\% \hspace{1cm}                                       \\
htru2             & 10022                     & 901                                     & 9121                                    & 9\% \hspace{1cm}                                        \\
letter            & 11200                     & 2167                                    & 9033                                    & 19\%  \hspace{1cm}                                      \\
mnist             & 48000                     & 4729                                    & 43271                                   & 10\%  \hspace{1cm}                                      \\
onlinenews        & 22200                     & 2899                                    & 19301                                   & 13\%  \hspace{1cm}                                      \\
pageblocks        & 3065                      & 118                                     & 2947                                    & 4\%  \hspace{1cm}                                       \\
redwine           & 895                       & 113                                     & 782                                     & 13\%  \hspace{1cm}                                      \\
sat               & 3548                      & 819                                     & 2729                                    & 23\% \hspace{1cm}                                       \\
sensorless        & 32765                     & 11934                                   & 20831                                   & 36\% \hspace{1cm}                                       \\
shuttle           & 34800                     & 7408                                    & 27392                                   & 21\%  \hspace{1cm}                                      \\
whitewine         & 2743                      & 587                                     & 2156                                    & 21\%  \hspace{1cm}                                      \\ \bottomrule
\end{tabular}
\end{table}

For the tabular datasets, we construct a multi-layer perceptron (MLP) with two hidden layers. Each layer has 100 nodes. For the image datasets, we construct a convolutional neural network (CNN) with two convolutional layers and two dense layers. In the training process, we use the standard gradient descent algorithm for both the BCE and AP-Perf networks. We use the learning rate of 0.01 for the BCE networks and 0.003 for the AP-Perf networks.
We select the learning rate values for both methods based on the training and validation test performance plot over 100 epochs.

For both methods, we perform a cross-validation using validation set to select the best L2 regularization among $\lambda = \{0, 0.001, 0.01, 0.1 \}$. 
After the training session finished, we compute the value of the metric for prediction in the testing dataset. 
For both methods, we select the predictive models that achieve the best metric in the validation set. We also implement an early stopping technique based on the validation set to avoid overfitting. Even though we run all the networks for 100 epochs, we select the parameters on the epoch that produce the best metric on the validation set. We then use this parameter to make predictions on the testing set. 

\section{Code Examples for Constructing Performance Metrics}

\label{sec:code-appendix}

\subsection{Commonly Used Performance Metrics}

Below are some code examples for constructing some of commonly used performance metrics. 

\begin{minted}[]{text}
@metric Accuracy      # Accuracy
function define(::Type{Accuracy}, C::ConfusionMatrix)
    return (C.tp + C.tn) / (C.all)  
end

accuracy_metric = Accuracy()

@metric Precision       # Precision
function define(::Type{Precision}, C::ConfusionMatrix)
    return C.tp / C.pp
end   

prec = Precision()
special_case_positive!(prec)

@metric Recall       # Recall / Sensitivity
function define(::Type{Recall}, C::ConfusionMatrix)
    return C.tp / C.ap
end   

rec = Recall()
special_case_positive!(rec)

@metric Specificity       # Specificity
function define(::Type{Specificity}, C::ConfusionMatrix)
    return C.tn / C.an
end   
spec = Specificity()
special_case_negative!(spec)

@metric F1Score         # F1 Score
function define(::Type{F1Score}, C::ConfusionMatrix)
    return (2 * C.tp) / (C.ap + C.pp)  
end   

f1_score = F1Score()
special_case_positive!(f1_score)

@metric GM_PrecRec      # Geometric Mean of Prec and Rec
function define(::Type{GM_PrecRec}, C::ConfusionMatrix)
    return C.tp / sqrt(C.ap * C.pp)  
end   

gpr = GM_PrecRec()
special_case_positive!(gpr)

@metric Informedness     # informedness
function define(::Type{Informedness}, C::ConfusionMatrix)
    return C.tp / C.ap + C.tn / C.an - 1
end   

inform = Informedness()
special_case_positive!(inform)
special_case_negative!(inform)

@metric Kappa       # Cohen's kappa score
function define(::Type{Kappa}, C::ConfusionMatrix)
    num = (C.tp + C.tn) / C.all - (C.ap * C.pp + C.an * C.pn) / C.all^2
    den = 1 - (C.ap * C.pp + C.an * C.pn) / C.all^2
    return num / den
end 

kappa = Kappa()
special_case_positive!(kappa)
special_case_negative!(kappa)

@metric PrecisionGvRecall         # precision given recall >= 0.8
function define(::Type{PrecisionGvRecall}, C::ConfusionMatrix)
    return C.tp / C.pp
end   
function constraint(::Type{PrecisionGvRecall}, C::ConfusionMatrix)
    return C.tp / C.ap >= 0.8
end   

precision_gv_recall = PrecisionGvRecall()
special_case_positive!(precision_gv_recall)
cs_special_case_positive!(precision_gv_recall, true)

@metric RecallGvPrecision         # recall given precision
function define(::Type{RecallGvPrecision}, C::ConfusionMatrix)
    return C.tp / C.pp
end   
function constraint(::Type{RecallGvPrecision}, C::ConfusionMatrix)
    return C.tp / C.ap >= 0.8
end   

recal_gv_precision = RecallGvPrecision()
special_case_positive!(recal_gv_precision)
cs_special_case_positive!(recal_gv_precision, true)

@metric PrecisionGvRecallSpecificity         # precision given recall >= 0.8 and specificity >= 0.8
function define(::Type{PrecisionGvRecallSpecificity}, C::ConfusionMatrix)
    return C.tp / C.pp
end   
function constraint(::Type{PrecisionGvRecallSpecificity}, C::ConfusionMatrix)
    return [C.tp / C.ap >= 0.8,
            C.tn / C.an >= 0.8]
end   

precision_gv_recall_spec = PrecisionGvRecallSpecificity()
special_case_positive!(precision_gv_recall_spec)
cs_special_case_positive!(precision_gv_recall_spec, [true, false])
cs_special_case_negative!(precision_gv_recall_spec, [false, true])

\end{minted}

\subsection{Performance Metrics with Arguments}

Our framework also supports writing performance metric with arguments, for example, the F$_\beta$ score metric which depends on the value of $\beta$. Below are some examples on constructing metrics with arguments.
\begin{minted}[]{text}
@metric FBeta beta          # F-Beta
function define(::Type{FBeta}, C::ConfusionMatrix, beta)
    return ((1 + beta^2) * C.tp) / (beta^2 * C.ap + C.pp)  
end   

f1_score = FBeta(1)
special_case_positive!(f1_score)

f2_score = FBeta(2)
special_case_positive!(f2_score)

# precision given recall
@metric PrecisionGvRecall th
function define(::Type{PrecisionGvRecall}, C::ConfusionMatrix, th)
    return C.tp / C.pp
end   

function constraint(::Type{PrecisionGvRecall}, C::ConfusionMatrix, th)
    return C.tp / C.ap >= th
end   

precision_gv_recall_80 = PrecisionGvRecall(0.8)
special_case_positive!(precision_gv_recall_80)
cs_special_case_positive!(precision_gv_recall_80, true)

precision_gv_recall_60 = PrecisionGvRecall(0.6)
special_case_positive!(precision_gv_recall_60)
cs_special_case_positive!(precision_gv_recall_60, true)

precision_gv_recall_95 = PrecisionGvRecall(0.95)
special_case_positive!(precision_gv_recall_95)
cs_special_case_positive!(precision_gv_recall_95, true)

@metric PrecisionGvRecallSpecificity th1 th2        # precision given recall >= th1 and specificity >= th2
function define(::Type{PrecisionGvRecallSpecificity}, C::ConfusionMatrix, th1, th2)
    return C.tp / C.pp
end   
function constraint(::Type{PrecisionGvRecallSpecificity}, C::ConfusionMatrix, th1, th2)
    return [C.tp / C.ap >= th1,
            C.tn / C.an >= th2]
end   

precision_gv_recall_spec = PrecisionGvRecallSpecificity(0.8, 0.8)
special_case_positive!(precision_gv_recall_spec)
cs_special_case_positive!(precision_gv_recall_spec, [true, false])
cs_special_case_negative!(precision_gv_recall_spec, [false, true])

\end{minted}

\section{Linear Program Solver using the ADMM Technique}
\label{sec:admm-appendix}

In this section we construct an ADMM formulation for solving the inner optimization over $\Qbf$ in Eq. \eqref{eq:dual-marginal}. The optimization can also be solved using any linear program solver as shown in the Appendix \ref{sec:thm-lp-appendix}. However, the runtime complexity of solving the LP is $O(m^6)$ where $m$ is the batch size, which makes it impractical for a batch of size greater than 30 samples. Our ADMM formulation reduces the runtime complexity to $O(m^3)$. 

We consider an extension of the family of evaluation metrics in Eq. \eqref{eq:metric} to also include the false positive and the false negative in the numerator of the fractions, i.e.,
\begin{equation}
    \text{metric}(\yvechat,\yvec) = \sum_j \frac{ a_j \tpz + b_j \tnz + c_j \fpz + d_j \fnz + f_j(\ppz,\apz)}{g_j(\ppz, \apz)},
\end{equation}
where $a_j$, $b_j$, $c_j$, and $d_j$ are constants.

\subsection{ADMM Formulation for Metrics with the Special Case for True Positive}

\label{sec:admm-true-pos}

We start with a task where the metric enforces a special case for true positive (for example, the precision, recall, and F1-score). In this task, the optimization over $\Qbf$ in Eq. \eqref{eq:dual-marginal} becomes:
\begin{align}
	& \min_{\Qbf \in \Delta}
     \max_{\Pbf \in \Delta}  \bigg[
      \sum_{k,l \in [1,n]} \sum_j \tfrac{1}{g_j(k, l)} \Big\{ a_j [ \pvec_k^1 \cdot \qvec_l^1] + b_j [ \pvec_k^0 \cdot \qvec_l^0] + c_j [ \pvec_k^1 \cdot \qvec_l^0] + d_j [ \pvec_k^0 \cdot \qvec_l^1] \label{eq:dual-marginal-admm}\\
    &  \qquad \qquad \qquad + f_j(k,l) r_k s_l \Big\} +  \Pcal(\zerovec) \Qcal(\zerovec) - \langle \Qbf^\intercal \onevec, \Psi^\intercal \theta \rangle \bigg]. \notag 
\end{align}

In this section we will use matrix notations in our formulation, extending our vector notations in Appendix \ref{sec:dual-marginal-appendix}. Using matrix notations, Eq. \eqref{eq:dual-marginal-admm} can be written as:
\begin{align}
    \min_{\{\Qbf_{1},\Qbf_{0},\svec,v_0\} \in \Delta}
     \max_{\{\Pbf_{1},\Pbf_{0},\rvec,u_0\} \in \Delta} \;
      &\left\langle \Mbf_1, \Pbf_1^\intercal \Qbf_1 \right\rangle
      + \left\langle \Mbf_2, \Pbf_1^\intercal \Qbf_0 \right\rangle
      + \left\langle \Mbf_3, \Pbf_0^\intercal \Qbf_1 \right\rangle \\
      & + \left\langle \Mbf_4, \Pbf_0^\intercal \Qbf_0 \right\rangle + \left\langle \Mbf_5, \rvec  \svec^\intercal \right\rangle + u_0 v_0  - \langle \Qbf_1, {\bf \Omega} \rangle, \notag  
\end{align}
where the matrix variables $\Qbf_1$, $\Qbf_0$, $\Pbf_1$, and $\Pbf_0$ represent:
\begin{align*}
    [\Qbf_1]_{i,j} &= \Qcal(\ychk_i = 1, \textstyle\sum_l \ychk_l = j), \quad i,j \in \{1,\hdots,n\} \\
    [\Qbf_0]_{i,j} &= \Qcal(\ychk_i = 0, \textstyle\sum_l \ychk_l = j), \quad i,j \in \{1,\hdots,n\}  \\
    [\Pbf_1]_{i,j} &= \Pcal(\yhat_i = 1, \textstyle\sum_l \yhat_l = j), \quad i,j \in \{1,\hdots,n\}  \\
    [\Pbf_0]_{i,j} &= \Pcal(\yhat_i = 0, \textstyle\sum_l \yhat_l = j), \quad i,j \in \{1,\hdots,n\},
\end{align*}
the vector and scalar variables represent:
\begin{align*}
    [\svec]_j &= \Qcal( \textstyle\sum_l \ychk_l = j), \quad j \in \{1,\hdots,n\}  \\
    v_0 &= \Qcal( \textstyle\sum_l \ychk_l = 0)\\
     [\rvec]_j &= \Pcal( \textstyle\sum_l \yhat_l = j), \quad j \in \{1,\hdots,n\} \\
    u_0 &= \Pcal( \textstyle\sum_l \yhat_l = 0),
\end{align*}
and the matrix ${\bf \Omega} = \Psi^\intercal \theta \onevec^\intercal$.

The matrix coefficients $\Mbf_1$, $\Mbf_2$, $\Mbf_3$, $\Mbf_4$, and $\Mbf_5$ are computed from the performance metric, where each cell $k,l$ of the matrices represents:
\begin{align*}
    [\Mbf_1]_{k,l} &= \sum_j \frac{a_j}{g_j(k,l)}, \quad
    [\Mbf_2]_{k,l} = \sum_j \frac{b_j}{g_j(k,l)}, \quad
    [\Mbf_3]_{k,l} = \sum_j \frac{c_j}{g_j(k,l)},  \\
    [\Mbf_4]_{k,l} &= \sum_j \frac{d_j}{g_j(k,l)}, \quad
    [\Mbf_5]_{k,l} = \sum_j \frac{f_j(k,l)}{g_j(k,l)}.
\end{align*}

We write the original marginal distribution constraint $\Delta$ over $\Pbf$ in matrix notations over $\{\Pbf_{1},\Pbf_{0},\rvec,u_0\}$ as:
\begin{align*}
    &\Pbf_1 \geq 0, \; \Pbf_0 \geq 0, \; \rvec \geq 0, \; u_0 \geq 0 \\
    &\rvec = \diag(\kappavec) \Pbf_1^\intercal \onevec \\
    &\rvec^\intercal \onevec + u_0 = 1 \\
    & \Pvec_1 + \Pvec_0 = \onevec \rvec^\intercal,
\end{align*}
where: $\kappavec = [\tfrac11,\tfrac12,\hdots, \tfrac{1}{n}]^\intercal$. All of the inequalities are element-wise.

Similarly, we write the original marginal distribution constraint $\Delta$ over $\Qbf$ in matrix notations over $\{\Qbf_{1},\Qbf_{0},\svec,v_0\}$ as:
\begin{align*}
    &\Qbf_1 \geq 0, \; \Qbf_0 \geq 0, \; \svec \geq 0, \; v_0 \geq 0 \\
    &\svec = \diag(\kappavec) \Qbf_1^\intercal \onevec \\
    &\svec^\intercal \onevec + v_0 = 1 \\
    & \Qvec_1 + \Qvec_0 = \onevec \svec^\intercal.
\end{align*}

\subsubsection{Simplification and Reformulation}

As mentioned in Appendix \ref{sec:dual-marginal-appendix}, we can compute all the variables for $\Pcal(y_i =0,\hdots)$
from the variables for $\Pcal(y_i =1,\hdots)$.
Specifically, we can derive $\Pbf_0$, $\rvec$, and $u_0$ from $\Pbf_1$. Let we denote $\Pbf = \Pbf_1$, then the equalities below hold:
\begin{align}
    \Pbf_0 &= \one \onevec^\intercal \Pbf \diag(\kappavec) - \Pbf \\
    \rvec &= \diag(\kappavec) \Pbf^\intercal \onevec  \\
    u_0 &= 1-\onevec^\intercal \Pbf \diag(\kappavec) \onevec,
\end{align}
and similarly for the adversary's variables, where $\Qbf = \Qbf_1$:
\begin{align}
    \Qbf_0 &= \one \onevec^\intercal \Qbf \diag(\kappavec) - \Qbf \\
    \svec &= \diag(\kappavec) \Qbf^\intercal \onevec  \\
    v_0 &= 1-\onevec^\intercal \Qbf \diag(\kappavec) \onevec.
\end{align}

Using this notation, we write Eq. \eqref{eq:dual-marginal-admm} as:
\begin{align}
    \min_{\Qbf \in \Delta}
     \max_{\Pbf \in \Delta} \;
      &\left\langle \Mbf_1, \Pbf^\intercal \Qbf \right\rangle
      + \left\langle \Mbf_2, \Pbf^\intercal (\onevec\onevec^\intercal \Qbf \diag(\kappavec) - \Qbf) \right\rangle
      + \left\langle \Mbf_3, (\one \onevec^\intercal \Pbf \diag(\kappavec) - \Pbf)^\intercal \Qbf \right\rangle  \label{eq:dual-marginal-admm-pq}
      \\
      & + \left\langle \Mbf_4, (\one \onevec^\intercal \Pbf \diag(\kappavec) - \Pbf)^\intercal (\onevec\onevec^\intercal \Qbf \diag(\kappavec) - \Qbf) \right\rangle  + \left\langle \Mbf_5,   \diag(\kappavec) \Pbf^\intercal \onevec \onevec^\intercal \Qbf \diag(\kappavec) \right\rangle \notag \\
      &+ (1-\onevec^\intercal \Pbf \diag(\kappavec) \onevec )(1 - \onevec^\intercal  \Qbf \diag(\kappavec) \onevec)  - \langle \Qbf, {\bf \Omega} \rangle \notag  
\end{align}
The constraint set $\Delta$ for $\Pbf$ is:
\begin{align*}
    &\Pbf \geq 0 \\
    &\onevec^\intercal \Pbf \diag(\kappavec) \onevec \leq 1 \\
    & \Pbf \leq \onevec\onevec^\intercal \Pbf \diag(\kappavec),
\end{align*}
and similarly for $\Qbf$:
\begin{align*}
    &\Qbf \geq 0 \\
    &\onevec^\intercal \Qbf \diag(\kappavec) \onevec \leq 1 \\
    & \Qbf \leq \onevec\onevec^\intercal \Qbf \diag(\kappavec),
\end{align*}
where all of the inequalities are element-wise.
This matrix inequalities for defining $\Delta$ is equivalent with the inequalities in Eq. \eqref{eq:delta}.

By rearranging the variables, we write Eq. \eqref{eq:dual-marginal-admm-pq} as:
\begin{align}
    &\min_{\Qbf \in \Delta}
     \max_{\Pbf \in \Delta} \;
      \left\langle \Pbf, \Qbf \Mbf_1^\intercal \right\rangle
      + \left\langle \Pbf, (\onevec\onevec^\intercal \Qbf \diag(\kappavec) - \Qbf) \Mbf_2^\intercal \right\rangle 
      + \left\langle \one \onevec^\intercal \Pbf \diag(\kappavec) - \Pbf, \Qbf \Mbf_3^\intercal \right\rangle \\
      & \qquad \qquad  + \left\langle \one \onevec^\intercal \Pbf \diag(\kappavec) - \Pbf,  (\onevec\onevec^\intercal \Qbf \diag(\kappavec) - \Qbf) \Mbf_4^\intercal \right\rangle  + \left\langle \Pbf, \onevec \onevec^\intercal \Qbf \diag(\kappavec) \Mbf_5^\intercal \diag(\kappavec) \right\rangle
        \notag
      \\
      & \qquad \qquad + \langle \Pbf, \onevec \onevec^\intercal \Qbf \diag(\kappavec) \onevec \onevec^\intercal \diag(\kappavec) \rangle - \langle \Pbf, \onevec \onevec^\intercal \diag(\kappavec) \rangle - \langle \Qbf, \onevec \onevec^\intercal \diag(\kappavec) \rangle + 1 - \langle \Qbf , {\bf \Omega} \rangle \notag  \phantom{\qquad\qquad\qquad\qquad} \\ \phantom{=} \notag 
\end{align}
\begin{align}
      =&\min_{\Qbf \in \Delta}
     \max_{\Pbf \in \Delta} \;
      1 - \langle \Qbf, \onevec \onevec^\intercal \diag(\kappavec) \rangle - \langle \Qbf, {\bf \Omega} \rangle \\
      &+ \left\langle \Pbf, \Qbf \Mbf_1^\intercal + (\onevec\onevec^\intercal \Qbf \diag(\kappavec) - \Qbf) \Mbf_2^\intercal + \onevec \onevec^\intercal \Qbf \diag(\kappavec) \Mbf_5^\intercal \diag(\kappavec) + \onevec \onevec^\intercal \Qbf \diag(\kappavec) \onevec \onevec^\intercal \diag(\kappavec) \rangle - \onevec \onevec^\intercal \diag(\kappavec) \right\rangle
       \notag \\
      &+ \left\langle \one \onevec^\intercal \Pbf \diag(\kappavec) - \Pbf , \Qbf \Mbf_3^\intercal + (\onevec\onevec^\intercal \Qbf \diag(\kappavec) - \Qbf ) \Mbf_4^\intercal \right\rangle  \notag   \\ \phantom{=} \notag \\
      =&\min_{\Qbf \in \Delta}
     \max_{\Pbf \in \Delta} \;
      1 - \langle \Qbf, \onevec \onevec^\intercal \diag(\kappavec) \rangle - \langle \Qbf, {\bf \Omega} \rangle \\
      &+ \left\langle \Pbf, \Qbf \Mbf_1^\intercal + (\onevec\onevec^\intercal \Qbf \diag(\kappavec) - \Qbf) \Mbf_2^\intercal + \onevec \onevec^\intercal \Qbf \diag(\kappavec) \Mbf_5^\intercal \diag(\kappavec) + \onevec \onevec^\intercal \Qbf \diag(\kappavec) \onevec \onevec^\intercal \diag(\kappavec) \rangle - \onevec \onevec^\intercal \diag(\kappavec) \right\rangle
       \notag \\
      &+ \left\langle \Pbf , \onevec\onevec^\intercal \Qbf \Mbf_3^\intercal \diag(\kappavec) + \onevec\onevec^\intercal \onevec\onevec^\intercal \Qbf \diag(\kappavec) \Mbf_4^\intercal \diag(\kappavec) - \onevec\onevec^\intercal \Qbf \Mbf_4^\intercal \diag(\kappavec) 
      - \Qbf \Mbf_3^\intercal - \onevec\onevec^\intercal \Qbf \diag(\kappavec) \Mbf_4^\intercal + \Qbf \Mbf_4^\intercal  
      \right\rangle  \notag
\end{align}

Given a fixed $\Qbf$ maximizing $\Pbf \in \Delta$ over a linear objective reduces to finding the column $k$ that has the maximum sum of $k$ largest elements in the column, with the additional restriction that it has to be greater than zero. We then simplify the formulation above as:
\begin{align}
    &\min_{\Qbf \in \Delta} f(\Abf \Qbf \Bbf + \Qbf \Cbf + \Dbf)
       + \langle \Qbf, \Ebf \rangle + c  \label{eq:admm-objective}
\end{align}
where:
\begin{align}
    f(\Xbf) =&\; \max(0, \max_k \; \text{sum-k-largest}( \Xbf_{(:,k)})) \\
    \Abf = &\; \onevec\onevec^\intercal \\
    \Bbf = &\; \diag(\kappavec) \Mbf_2^\intercal + \diag(\kappavec) \Mbf_5^\intercal \diag(\kappavec) + \diag(\kappavec) \onevec \onevec^\intercal \diag(\kappavec) \\
    &\quad + \Mbf_3^\intercal \diag(\kappavec) + n \diag(\kappavec) \Mbf_4^\intercal \diag(\kappavec) - \Mbf_4^\intercal \diag(\kappavec) - \diag(\kappavec) \Mbf_4^\intercal \notag \\
    \Cbf = &\; \Mbf_1^\intercal - \Mbf_2^\intercal - \Mbf_3^\intercal +\Mbf_4^\intercal \\
    \Dbf = &\; - \onevec \onevec^\intercal \diag(\kappavec) \\
    \Ebf = &\; - \onevec \onevec^\intercal \diag(\kappavec) - {\bf \Omega } \\
    c = & \;1
\end{align}

\subsubsection{ADMM Formulation}

We perform an alternating direction method of multipliers (ADMM) optimization to optimize Eq. \eqref{eq:admm-objective}. We split the optimization into three variables: $\Qbf, \Xbf,$ and $\Zbf$.
\begin{align}
    \min_{\Qbf, \Xbf, \Zbf}& \; f(\Zbf)
       + \langle \Qbf, \Ebf \rangle + \Ibf_{\Delta}(\Qbf) + c \\
    \text{s.t. } & \Zbf = \Abf \Xbf \Bbf + \Xbf \Cbf + \Dbf \notag \\
    & \Qbf = \Xbf \notag,
\end{align}
where $\Ibf_{\Delta}(\Qbf)$ returns 0 if $\Qbf \in \Delta$ or $\infty$ otherwise.

The augmented Lagrangian (scaled version) for this optimization is:
\begin{align}
    &\Lcal(\Qbf, \Xbf, \Zbf, \Ubf, \Wbf) = \notag \\
    & \qquad f(\Zbf)
       + \langle \Qbf, \Ebf \rangle + \Ibf_{\Delta}(\Qbf) + c + \frac{\rho}{2} \| \Abf \Xbf \Bbf + \Xbf \Cbf + \Dbf - \Zbf + \Ubf \|_F^2 + \frac{\rho}{2} \| \Xbf - \Qbf + \Wbf \|_F^2,
\end{align}
where $\|\cdot\|_F$ denotes the Frobenius  norm of a matrix, $\rho$ is the ADMM penalty parameter, whereas  $\Ubf$ and $\Wbf$ are the dual variables for the constraint $\Zbf = \Abf \Xbf \Bbf + \Xbf \Cbf + \Dbf$ and $\Qbf = \Xbf$ respectively.

The ADMM updates for each variable are explained below:
\begin{enumerate}[wide=0pt]

\item Update for $\Qbf$: a projection operation
\begin{align}
    \Qbf^{(t+1)} &= \argmin_{\Qbf} \left\{ \langle \Qbf, \Ebf \rangle + \Ibf_{\Delta}(\Qbf) + \frac{\rho}{2} \| \Xbf^{(t)} - \Qbf + \Wbf^{(t)} \|_F^2 \right\} \\
    &= \argmin_{\Qbf \in \Delta}  \frac{1}{2} \| \tfrac{1}{\rho} (\rho (\Xbf^{(t)} + \Wbf^{(t)}) - \Ebf ) - \Qbf \|_F^2 \\
    &= \text{Proj}_{\Delta} (\tfrac{1}{\rho} (\rho (\Xbf^{(t)} + \Wbf^{(t)}) - \Ebf )) \label{eq:admm-proj-q}
\end{align}

\item Update for $\Zbf$: a proximal operation.
\begin{align}
    \Zbf^{(t+1)} &= \argmin_{\Zbf} \left\{ f(\Zbf) + \frac{\rho}{2} \| \Abf \Xbf^{(t)} \Bbf + \Xbf^{(t)} \Cbf + \Dbf - \Zbf + \Ubf^{(t)} \|_F^2 \right\} \\
    &= \text{prox}_{f, 1/\rho} (\Abf \Xbf^{(t)} \Bbf + \Xbf^{(t)} \Cbf + \Dbf + \Ubf^{(t)}) \label{eq:admm-prox}
\end{align}

\item Update for $\Xbf$: Sylvester equation
\begin{align}
    \Xbf^{(t+1)} &= \argmin_{\Xbf} \left\{ \frac{\rho}{2} \| \Abf \Xbf \Bbf + \Xbf \Cbf + \Dbf - \Zbf^{(t+1)} + \Ubf^{(t)} \|_F^2 + \frac{\rho}{2} \| \Xbf - \Qbf^{(t+1)} + \Wbf^{(t)} \|_F^2 \right\} \\
    &= \argmin_{\Xbf} \left\{ \frac{1}{2} \| \Abf \Xbf \Bbf + \Xbf \Cbf + \Dbf - \Zbf^{(t+1)} + \Ubf^{(t)} \|_F^2 + \frac{1}{2} \| \Xbf - \Qbf^{(t+1)} + \Wbf^{(t)} \|_F^2 \right\}
\end{align}

We solve the minimization above by setting the gradient  w.r.t. $\Xbf$ to zero. Removing the superscript over iteration $t$, the gradient of the objective above w.r.t. $\Xbf$ is:
\begin{align}
\Abf^\intercal \Abf \Xbf \Bbf \Bbf^\intercal + \Abf^\intercal \Xbf \Cbf \Bbf^\intercal + \Abf \Xbf \Bbf \Cbf^\intercal + \Xbf \Cbf \Cbf^\intercal + \Abf^\intercal (\Dbf - \Zbf + \Ubf) \Bbf^\intercal + (\Dbf - \Zbf + \Ubf) \Cbf^\intercal + \Xbf  + \Wbf - \Qbf.
\end{align}
Since $\Abf = \onevec\onevec^\intercal$, the gradient can be simplified as:
\begin{align}
&\Abf \Xbf n \Bbf \Bbf^\intercal + \Abf \Xbf \Cbf \Bbf^\intercal + \Abf \Xbf \Bbf \Cbf^\intercal + \Xbf \Cbf \Cbf^\intercal + \Abf (\Dbf - \Zbf + \Ubf) \Bbf^\intercal + (\Dbf - \Zbf + \Ubf) \Cbf^\intercal + \Xbf + \Wbf - \Qbf \\
=\;&  \Abf \Xbf (n \Bbf \Bbf^\intercal + \Cbf \Bbf^\intercal + \Bbf \Cbf^\intercal ) + \Xbf ( \Cbf \Cbf^\intercal + \Ibf ) + \Abf (\Dbf - \Zbf + \Ubf) \Bbf^\intercal + (\Dbf - \Zbf + \Ubf) \Cbf^\intercal  +  \Wbf - \Qbf.
\end{align}
Let $\Fbf =  \Abf (\Dbf - \Zbf + \Ubf) \Bbf^\intercal + (\Dbf - \Zbf + \Ubf) \Cbf^\intercal + \Wbf - \Qbf $. The optimal $\Xbf$ can be found by solving a Sylvester equation below:
\begin{align}
    \Abf \Xbf ( n \Bbf \Bbf^\intercal + \Cbf \Bbf^\intercal + \Bbf \Cbf^\intercal ) + \Xbf ( \Cbf \Cbf^\intercal + \Ibf ) + \Fbf &= 0 \\
    \Abf \Xbf ( n \Bbf \Bbf^\intercal + \Cbf \Bbf^\intercal + \Bbf \Cbf^\intercal ) + \Xbf ( \Cbf \Cbf^\intercal + \Ibf ) &= - \Fbf \\
    \Abf \Xbf ( n \Bbf \Bbf^\intercal + \Cbf \Bbf^\intercal + \Bbf \Cbf^\intercal  ) (  \Cbf \Cbf^\intercal +  \Ibf )^{-1}  + \Xbf &= -\Fbf (  \Cbf \Cbf^\intercal +  \Ibf )^{-1}. \label{eq:admm-sylvester}
\end{align}
Note that a Sylvester equation is a matrix equation in the form of $A X B + X = C$ or $A X + X B = C$. 

\item Update for $\Ubf$:
\begin{align}
    \Ubf^{(t+1)} = \Ubf^{(t)} + \Abf \Xbf^{(t)} \Bbf + 
\Xbf^{(t)} \Cbf + \Dbf - \Zbf^{(t+1)}.
\end{align}
\item Update for $\Wbf$:
\begin{align}
    \Wbf^{(t+1)} = \Wbf^{(t)} + \Xbf^{(t+1)} - \Qbf^{(t+1)}.
\end{align}

Please go to Section \ref{sec:projection-appendix},  \ref{sec:proximal-appendix}, and  \ref{sec:sylvester-appendix} for the detailed algorithms for the projection, proximal operator, and Sylvester equation solver.

\end{enumerate}

\subsection{ADMM Formulation for Metrics without Special Cases}

For the metric that does not enforce any special cases, the optimization over $\Qbf$ is:
\begin{align}
	& \min_{\Qbf \in \Delta}
     \max_{\Pbf \in \Delta}  \bigg[
      \sum_{k,l \in [0,n]} \sum_j \tfrac{1}{g_j(k, l)} \Big\{ a_j [ \pvec_k^1 \cdot \qvec_l^1] + b_j [ \pvec_k^0 \cdot \qvec_l^0] + c_j [ \pvec_k^1 \cdot \qvec_l^0] + d_j [ \pvec_k^0 \cdot \qvec_l^1] + f_j(k,l) r_k s_l \Big\}  - \langle \Qbf^\intercal \onevec, \Psi^\intercal \theta \rangle \bigg].
\end{align}

Since the summation index in the equation above is from 0 to $n$, whereas our variables $\Pbf$ and $\Qbf$ represent the indices from 1 to $n$, we need to treat the summation over index 0 separately. Specifically, the matrix notation optimization is now:
\begin{align}
    \min_{\{\Qbf_{1},\Qbf_{0},\svec,v_0\} \in \Delta}
     \max_{\{\Pbf_{1},\Pbf_{0},\rvec,u_0\} \in \Delta} \;
      &\left\langle \Mbf_1, \Pbf_1^\intercal \Qbf_1 \right\rangle
      + \left\langle \Mbf_2, \Pbf_1^\intercal \Qbf_0 \right\rangle
      + \left\langle \Mbf_3, \Pbf_0^\intercal \Qbf_1 \right\rangle  + \left\langle \Mbf_4, \Pbf_0^\intercal \Qbf_0 \right\rangle + \left\langle \Mbf_5, \rvec  \svec^\intercal \right\rangle  \\
      & + m_{4[0,0]} u_0 v_0 + \langle \mvec_{4[0,:]}, u_0 \onevec^\intercal \Qbf_0 \rangle + \langle \mvec_{4[:,0]}, \Pbf_0^\intercal \onevec v_0 \rangle \notag  \\
      & + m_{5[0,0]} u_0 v_0 + \langle \mvec_{5[0,:]}, u_0 \svec^\intercal \rangle + \langle \mvec_{5[:,0]}, \rvec v_0 \rangle - \langle \Qbf_1, {\bf \Omega} \rangle, \notag
\end{align}
where:
\begin{align*}
    m_{4[0,0]} &= \sum_j \frac{d_j}{g_j(0,0)}, \quad
    \mvec_{4[0,l]} = \sum_j \frac{d_j}{g_j(0,l)}, \quad
    \mvec_{4[k,0]} = \sum_j \frac{d_j}{g_j(k,0)},  \\
    m_{5[0,0]} &= \sum_j \frac{f_j(0,0)}{g_j(0,0)}, \quad
    \mvec_{5[0,l]} = \sum_j \frac{f_j(0,l)}{g_j(0,l)}, \quad
    \mvec_{5[k,0]} = \sum_j \frac{f_j(k,0)}{g_j(k,0)}.
\end{align*}

Using the same technique as in Appendix \ref{sec:admm-true-pos}, we write the optimization over the matrix $\Pbf$ and $\Qbf$ only, and regroup the variables as follows:
\begin{align}
    &\min_{\Qbf \in \Delta}
     \max_{\Pbf \in \Delta} \;
      \left\langle \Mbf_1, \Pbf^\intercal \Qbf \right\rangle
      + \left\langle \Mbf_2, \Pbf^\intercal (\onevec\onevec^\intercal \Qbf \diag(\kappavec) - \Qbf) \right\rangle
      + \left\langle \Mbf_3, (\one \onevec^\intercal \Pbf \diag(\kappavec) - \Pbf)^\intercal \Qbf \right\rangle  
      \\
      & \qquad + \left\langle \Mbf_4, (\one \onevec^\intercal \Pbf \diag(\kappavec) - \Pbf)^\intercal (\onevec\onevec^\intercal \Qbf \diag(\kappavec) - \Qbf) \right\rangle  + \left\langle \Mbf_5,   \diag(\kappavec) \Pbf^\intercal \onevec \onevec^\intercal \Qbf \diag(\kappavec) \right\rangle \notag \\
      & \qquad+ (m_{4[0,0]} + m_{5[0,0]}) (1-\onevec^\intercal \Pbf \diag(\kappavec) \onevec )(1 - \onevec^\intercal  \Qbf \diag(\kappavec) \onevec)  \notag  \\
       & \qquad + \langle \mvec_{4[0,:]}, (1-\onevec^\intercal \Pbf \diag(\kappavec) \onevec) \onevec^\intercal (\onevec\onevec^\intercal \Qbf \diag(\kappavec) - \Qbf) \rangle + \langle \mvec_{4[:,0]}, (\one \onevec^\intercal \Pbf \diag(\kappavec) - \Pbf)^\intercal \onevec (1 - \onevec^\intercal  \Qbf \diag(\kappavec) \onevec) \rangle \notag  \\
       & \qquad +  \langle \mvec_{5[0,:]}, (1-\onevec^\intercal \Pbf \diag(\kappavec) \onevec) \onevec^\intercal \Qbf \diag(\kappavec) \rangle + \langle \mvec_{5[:,0]},  \diag(\kappavec) \Pbf^\intercal \onevec (1 - \onevec^\intercal  \Qbf \diag(\kappavec) \onevec) \rangle - \langle \Qbf, {\bf \Omega} \rangle \notag  \\ \phantom{=} \notag \\
       =\;& 
    \min_{\Qbf \in \Delta}
     \max_{\Pbf \in \Delta} \;
      m_{4[0,0]} + m_{5[0,0]} - \langle \Qbf, \onevec \onevec^\intercal \diag(\kappavec) (m_{4[0,0]} + m_{5[0,0]}) \rangle - \langle \Qbf, {\bf \Omega} \rangle \\
      &\qquad+ \left\langle \Qbf, n \onevec {\mvec_{4[0,:]}} \diag(\kappavec) - \onevec  {\mvec_{4[0,:]}} + \onevec  {\mvec_{5[0,:]}} \diag(\kappavec) \right\rangle \notag \\
      &\qquad+ \Big\langle \Pbf, \Big\{ \Qbf \Mbf_1^\intercal + (\onevec\onevec^\intercal \Qbf \diag(\kappavec) - \Qbf) \Mbf_2^\intercal + \onevec \onevec^\intercal \Qbf \diag(\kappavec) \Mbf_5^\intercal \diag(\kappavec) + \onevec \onevec^\intercal \Qbf \diag(\kappavec) \onevec \onevec^\intercal \diag(\kappavec) (m_{4[0,0]} + m_{5[0,0]}) \rangle \notag \\
      &\qquad \qquad - \onevec \onevec^\intercal \diag(\kappavec) (m_{4[0,0]} + m_{5[0,0]}) -  n \onevec \onevec^\intercal \Qbf \diag(\kappavec) \mvec_{4[0,:]}^\intercal \onevec^\intercal \diag(\kappavec) +  \onevec \onevec^\intercal \Qbf \mvec_{4[0,:]}^\intercal \onevec^\intercal \diag(\kappavec)  \notag \\
      &\qquad \qquad -  \onevec \onevec^\intercal \Qbf \diag(\kappavec) \mvec_{5[0,:]}^\intercal \onevec^\intercal \diag(\kappavec) + \onevec \mvec_{5[:,0]}^\intercal \diag(\kappavec) + \onevec \onevec^\intercal \Qbf \diag(\kappavec) \onevec \mvec_{5[:,0]}^\intercal \diag(\kappavec)
      \Big\} \Big\rangle
       \notag \\
      &\qquad+ \left\langle \one \onevec^\intercal \Pbf \diag(\kappavec) - \Pbf , \left\{ \Qbf \Mbf_3^\intercal + (\onevec\onevec^\intercal \Qbf \diag(\kappavec) - \Qbf ) \Mbf_4^\intercal +  \onevec \mvec_{4[:,0]}^\intercal - \onevec \onevec^\intercal \Qbf \diag(\kappavec) \onevec \mvec_{4[:,0]}^\intercal  \right\} \right\rangle  \notag 
\end{align}
\begin{align}
      =\;  &\min_{\Qbf \in \Delta}
     \max_{\Pbf \in \Delta} \;
      m_{4[0,0]} + m_{5[0,0]} + \left\langle \Qbf, \left\{ n \onevec {\mvec_{4[0,:]}} \diag(\kappavec) - \onevec  {\mvec_{4[0,:]}} + \onevec  {\mvec_{5[0,:]}} \diag(\kappavec) - \onevec \onevec^\intercal \diag(\kappavec) (m_{4[0,0]} + m_{5[0,0]}) - {\bf \Omega} \right\} \right\rangle \notag \\
      &\qquad+ \Big\langle \Pbf, \Big\{ \Qbf \Mbf_1^\intercal + (\onevec\onevec^\intercal \Qbf \diag(\kappavec) - \Qbf) \Mbf_2^\intercal + \onevec \onevec^\intercal \Qbf \diag(\kappavec) \Mbf_5^\intercal \diag(\kappavec) + \onevec \onevec^\intercal \Qbf \diag(\kappavec) \onevec \onevec^\intercal \diag(\kappavec) (m_{4[0,0]} + m_{5[0,0]}) \rangle \notag \\
      &\qquad \qquad - \onevec \onevec^\intercal \diag(\kappavec) (m_{4[0,0]} + m_{5[0,0]}) -  n \onevec \onevec^\intercal \Qbf \diag(\kappavec) \mvec_{4[0,:]}^\intercal \onevec^\intercal \diag(\kappavec) +  \onevec \onevec^\intercal \Qbf \mvec_{4[0,:]}^\intercal \onevec^\intercal \diag(\kappavec)  \notag \\
      &\qquad \qquad -  \onevec \onevec^\intercal \Qbf \diag(\kappavec) \mvec_{5[0,:]}^\intercal \onevec^\intercal \diag(\kappavec) + \onevec \mvec_{5[:,0]}^\intercal \diag(\kappavec) + \onevec \onevec^\intercal \Qbf \diag(\kappavec) \onevec \mvec_{5[:,0]}^\intercal \diag(\kappavec)
      \Big\} \Big\rangle
       \notag \\
      &\qquad + \Big\langle \Pbf , \Big\{ \onevec\onevec^\intercal \Qbf \Mbf_3^\intercal \diag(\kappavec) + n \onevec\onevec^\intercal \Qbf \diag(\kappavec) \Mbf_4^\intercal \diag(\kappavec) - \onevec\onevec^\intercal \Qbf \Mbf_4^\intercal \diag(\kappavec)  +
      n \onevec \mvec_{4[:,0]}^\intercal \diag(\kappavec) \notag \\
      & \qquad \qquad- n \onevec \onevec^\intercal \Qbf \diag(\kappavec) \onevec \mvec_{4[:,0]}^\intercal  \diag(\kappavec) 
      - \Qbf \Mbf_3^\intercal - \onevec\onevec^\intercal \Qbf \diag(\kappavec) \Mbf_4^\intercal + \Qbf \Mbf_4^\intercal - \onevec \mvec_{4[:,0]}^\intercal + \onevec \onevec^\intercal \Qbf \diag(\kappavec) \onevec \mvec_{4[:,0]}^\intercal \Big\}
      \Big\rangle 
\end{align}

As in Appendix \ref{sec:admm-true-pos}, the equation above can be simplified as:
\begin{align}
    &\min_{\Qbf \in \Delta} f(\Abf \Qbf \Bbf + \Qbf \Cbf + \Dbf)
       + \langle \Qbf, \Ebf \rangle + c 
\end{align}
where:
\begin{align}
    f(\Xbf) =&\; \max(0, \max_k \; \text{sum-k-largest}( \Xbf_{(:,k)})) \\
    \Abf = &\; \onevec\onevec^\intercal \\
    \Bbf = &\; \diag(\kappavec) \Mbf_2^\intercal + \diag(\kappavec) \Mbf_5^\intercal \diag(\kappavec) + \diag(\kappavec) \onevec \onevec^\intercal \diag(\kappavec) (m_{4[0,0]} + m_{5[0,0]}) \\
    & -  n \diag(\kappavec) \mvec_{4[0,:]}^\intercal \onevec^\intercal \diag(\kappavec) +  \mvec_{4[0,:]}^\intercal \onevec^\intercal \diag(\kappavec) -  \diag(\kappavec) \mvec_{5[0,:]}^\intercal \onevec^\intercal \diag(\kappavec) +  \diag(\kappavec) \onevec \mvec_{5[:,0]}^\intercal \diag(\kappavec) \notag \\
    & + \Mbf_3^\intercal \diag(\kappavec) + n \diag(\kappavec) \Mbf_4^\intercal \diag(\kappavec) - \Mbf_4^\intercal \diag(\kappavec) - n \diag(\kappavec) \onevec \mvec_{4[:,0]}^\intercal  \diag(\kappavec)  - \diag(\kappavec) \Mbf_4^\intercal \notag + \diag(\kappavec) \onevec \mvec_{4[:,0]}^\intercal \\
    \Cbf = &\; \Mbf_1^\intercal - \Mbf_2^\intercal - \Mbf_3^\intercal +\Mbf_4^\intercal \\
    \Dbf = &\; - \onevec \onevec^\intercal \diag(\kappavec) (m_{4[0,0]} + m_{5[0,0]}) + \onevec \mvec_{5[:,0]}^\intercal \diag(\kappavec) +
      n \onevec \mvec_{4[:,0]}^\intercal \diag(\kappavec) - \onevec \mvec_{4[:,0]}^\intercal \\
    \Ebf = &\; n \onevec {\mvec_{4[0,:]}} \diag(\kappavec) - \onevec  {\mvec_{4[0,:]}} + \onevec  {\mvec_{5[0,:]}} \diag(\kappavec) - \onevec \onevec^\intercal \diag(\kappavec) (m_{4[0,0]} + m_{5[0,0]}) - {\bf \Omega} \\
    c = & \; m_{4[0,0]} + m_{5[0,0]}
\end{align}

Since the form of the objective above is similar to the one in Appendix \ref{sec:admm-true-pos}, we use the same ADMM technique to solve the optimization over $\Qbf$. Note that only the constant variables that are defined by the form of the metric ($\Abf, \Bbf, \Cbf, \Dbf, \Ebf,$ and $c$) are modified from Eq. \eqref{eq:admm-objective}. All the ADMM updates remain the same.

\subsection{ADMM Formulation for Metrics with Special Case for True Negative}

For the metrics that enforce special cases for true negative only (e.g., specificity) or special cases for both true negative and true positive (e.g., the MCC and Kappa score), we use the optimization schemes for the metrics that do not enforce special cases for true negative, with a little modification. Specifically, we modify the coefficient matrix $\Mbf_1$ and $\Mbf_5$ by setting the  values in the $n$-th row and the $n$-th column to be zero, except for the $(n,n)$-th cell where we set it to one. Therefore, for the metrics that enforce special cases for both true positive and true negative, we have:
\begin{align}
    \min_{\{\Qbf_{1},\Qbf_{0},\svec,v_0\} \in \Delta}
     \max_{\{\Pbf_{1},\Pbf_{0},\rvec,u_0\} \in \Delta} \;
      &\left\langle \Mbf_1^\diamond, \Pbf_1^\intercal \Qbf_1 \right\rangle
      + \left\langle \Mbf_2, \Pbf_1^\intercal \Qbf_0 \right\rangle
      + \left\langle \Mbf_3, \Pbf_0^\intercal \Qbf_1 \right\rangle \\
      & + \left\langle \Mbf_4, \Pbf_0^\intercal \Qbf_0 \right\rangle + \left\langle \Mbf_5^\diamond, \rvec  \svec^\intercal \right\rangle + u_0 v_0  - \langle \Qbf_1, \Psi \rangle, \notag  
\end{align}
whereas for the metrics that enforce special cases for true negative only we have:
\begin{align}
    \min_{\{\Qbf_{1},\Qbf_{0},\svec,v_0\} \in \Delta}
     \max_{\{\Pbf_{1},\Pbf_{0},\rvec,u_0\} \in \Delta} \;
      &\left\langle \Mbf_1^\diamond, \Pbf_1^\intercal \Qbf_1 \right\rangle
      + \left\langle \Mbf_2, \Pbf_1^\intercal \Qbf_0 \right\rangle
      + \left\langle \Mbf_3, \Pbf_0^\intercal \Qbf_1 \right\rangle  + \left\langle \Mbf_4, \Pbf_0^\intercal \Qbf_0 \right\rangle + \left\langle \Mbf_5^\diamond, \rvec  \svec^\intercal \right\rangle  \\
      & + m_{4[0,0]} u_0 v_0 + \langle \mvec_{4[0,:]}, u_0 \onevec^\intercal \Qbf_0 \rangle + \langle \mvec_{4[:,0]}, \Pbf_0^\intercal \onevec v_0 \rangle \notag  \\
      & + m_{5[0,0]} u_0 v_0 + \langle \mvec_{5[0,:]}, u_0 \svec^\intercal \rangle + \langle \mvec_{5[:,0]}, \rvec v_0 \rangle - \langle \Qbf_1, \Psi \rangle, \notag
\end{align}
where:
\begin{align}
    \Mbf^\diamond_{i,j} = 
    \begin{cases}
        1 & \text{if } i = j = n \\
        0 & \text{if } (i = n \wedge j \neq n) \vee (i \neq n \wedge j = n) \\
        \Mbf_{i,j} & \text{otherwise.}
    \end{cases}
\end{align}
All other ADMM optimization techniques remain the same.

\subsection{Projection onto the Valid Marginal Probability Set}

\label{sec:projection-appendix}

In the ADMM updates for $\Qbf$ (Eq. \eqref{eq:admm-proj-q}), we need to perform a projection onto the set of valid marginal distributions $\Delta$. In this subsection, we will derive an algorithm to efficiently perform the projection.

Given a matrix $\Abf$ that is not necessary in the set $\Delta$, we want to find $\Pbf \in \Delta$ that minimizes the Euclidean distance between $\Abf$ and $\Pbf \in \Delta$. Specifically, we need to solve:
\begin{align}
    \min_{\Pbf \in \Delta} \tfrac12 \| \Pbf - \Abf \|_F^2.
\end{align}
In our vector notation (see. Appendix \ref{sec:dual-marginal-appendix}), this is equal to:
\begin{align}
     \min_{\{\pvec_k\}} & \; \tfrac12 \sum_k \| \pvec_k - \avec_k \|_2^2 \\
    \text{subject to: } & p_{i,k} \geq 0, \quad \forall i,k \in [1,n] \notag \\
    & p_{i,k} \leq \tfrac{1}{k} \textstyle\sum_j p_{j,k}, \quad \forall i,k \in [1,n] \notag \\
    & \textstyle\sum_k \tfrac{1}{k} \sum_i p_{i,k} \leq 1 \notag,
\end{align}
where $\pvec_k$ and $\avec_k$ are the $k$-th column of the $\Pbf$ and $\Abf$ respectively.

The constraints above can be written as:
\begin{align}
     \min_{\{\pvec_k \in \Cbb_k\}} & \; \tfrac12 \sum_k \| \pvec_k - \avec_k \|_2^2, \; \text{s.t.}  \textstyle\sum_k \frac{\pvec_{k}^\intercal \onevec}{k} \leq 1 \\
    \text{where: } & \Cbb_k = \{\pvec_k \mid \pvec_{k} \in [0, r_k]^n; \; r_k \geq 0; \; r_k = \tfrac{\pvec_{k}^\intercal \onevec}{k} \} \notag. 
\end{align}

Using the Lagrange multiplier technique, we write the dual optimization as:
\begin{align}
    & \max_{\eta \geq 0}  \min_{\{\pvec_k \in \Cbb_k\}} \; \tfrac12 \sum_k \| \pvec_k - \avec_k \|_2^2, + \eta \left( \textstyle\sum_k \frac{\pvec_{k}^\intercal \onevec}{k} - 1 \right) \\
    =& \max_{\eta \geq 0} - \eta +  \; \sum_k \min_{\pvec_k \in \Cbb_k} \left\{ \tfrac12 \| \pvec_k - \avec_k \|_2^2 + \tfrac{\eta}{k}  \pvec_{k}^\intercal \onevec \right\} \label{eq:obj_eta}
\end{align}

Given $\eta$, the inner minimization is now decomposable into each individual $\pvec_k$. 
For convenience, we drop the subscript $k$ in the next analysis, i.e.,
\begin{align}
    & \min_{\pvec \in \Cbb} \left\{ \tfrac12 \| \pvec - \avec \|_2^2 + \tfrac{\eta}{k}  \pvec^\intercal \onevec \right\}. \label{eq:solve_p_given_ak} \\
    \text{where: } & \Cbb = \{\pvec \mid \pvec \in [0, r]^n; \; r \geq 0; \; r = \tfrac{\pvec^\intercal \onevec}{k} \} \notag. 
\end{align}

This minimization problem admits a search-based analytical solution. 
We start with the $\bar{\pvec} = \avec - \frac{\eta}{k}$, which is the minimizer of the objective without the constraint as the proposed solution, and start with $r =\frac{\bar{\pvec}^\intercal}{k}$. If all of $p_i$ lies in $[0,r]$, we accept $\bar{\pvec}$ as the solution, otherwise, we iteratively reduce the value of the highest probability values in $\bar{\pvec}$, which automatically reduce the value of $r =\frac{\bar{\pvec}^\intercal}{k}$, and simultaneously setting negative values in $\bar{\pvec}$ as zero. This requires sorting the values in $\bar{\pvec}$ in a decreasing order.

Given we have the solution of Eq. \eqref{eq:solve_p_given_ak} for each column, we calculate the objective and gradient of Eq. \eqref{eq:solve_p_given_ak} with respect to $\eta$. Since it is just a one-dimensional optimization, we  efficiently solve it with a gradient-based optimization with box constraint of $\eta \geq 0$. Note that the objective is concave with respect to $\eta$.

\subsection{Proximal Operator for the ADMM Updates}

\label{sec:proximal-appendix}

In the ADMM updates for $\Zbf$ (Eq. \eqref{eq:admm-prox}), we need to perform a proximal operator for the function $f(\Xbf)$, i.e.:
\begin{align}
    f(\Xbf) =&\; \max(0, \max_k \; \text{sum-k-largest}( \Xbf_{(:,k)})).
\end{align}
The proximal operator over $f$ is:
\begin{align}
    \text{prox}_{f, 1/\rho} (\Xbf) &= \argmin_{\Zbf} \left\{ f(\Zbf) + \frac{\rho}{2} \| \Xbf - \Zbf \|_F^2 \right\} 
\end{align}
Note that $f(\Zbf)$ can be expanded as:
\begin{align}
       f(\Zbf) &= \max_{\Pbf \in \Delta} \; \langle \Pbf, \Zbf \rangle 
       =\min_{\Pbf \in \Delta} \; \langle \Pbf, -\Zbf \rangle =\min_{\Pbf} \left( \Ibf_{\Delta}(\Pbf) - \langle \Pbf, \Zbf \rangle \right) = \sup_{\Pbf} \left( \langle \Pbf, \Zbf \rangle - \Ibf_{\Delta}(\Pbf) \right) = \Ibf_{\Delta}^{*}(\Zbf) ,
\end{align}
where $\Ibf_{\Delta}^{*}(\Zbf)$ denotes the conjugate function of $\Ibf_{\Delta}(\Zbf) $. 

Based on Moreau Decomposition \citep{moreau1962decomposition}, we know that:
\begin{align}
    \text{prox}_{f} (\Xbf) &= \Xbf - \text{prox}_{\Ibf_{\Delta}} (\Xbf) \\
    & = \Xbf -  \argmin_{\Zbf} \left\{ \Ibf_{\Delta}(\Zbf) + \tfrac{1}{2} \| \Xbf - \Zbf \|_F^2 \right\}  \\
    & =\Xbf -  \argmin_{\Zbf \in \Delta} \tfrac{1}{2} \| \Xbf - \Zbf \|_F^2 \\
    &= \Xbf - \text{Proj}_{\Delta}(\Xbf)
\end{align}
Therefore, we can compute $\text{prox}_{f, 1/\rho} (\Xbf)$ as:
\begin{align}
    \text{prox}_{f, 1/\rho} (\Xbf) &= \Xbf -  \tfrac{1}{\rho} \; \text{prox}_{\rho f^*} (\rho \Xbf ) \\
    &= \Xbf -  \tfrac{1}{\rho} \; \text{Proj}_{\Delta}  (\rho \Xbf )
\end{align}

\subsection{Solving the Sylvester Equation in the ADMM update}

\label{sec:sylvester-appendix}

In the ADMM updates for $\Xbf$ (Eq. \eqref{eq:admm-sylvester}), we need solve a Sylvester equation in the form of:
\begin{align}
    \Abf \Xbf ( n \Bbf \Bbf^\intercal + \Cbf \Bbf^\intercal + \Bbf \Cbf^\intercal  ) (  \Cbf \Cbf^\intercal +  \Ibf )^{-1}  + \Xbf &= -\Fbf (  \Cbf \Cbf^\intercal +  \Ibf )^{-1}. 
\end{align}
Many linear algebra packages in most of program languages have the capability to solve a Sylvester equation. However, since our formulation contains matrices with special property, we develop a faster customized solver that utilizes the eigen-decomposition technique and exploits the fact that $\Abf$, $( n \Bbf \Bbf^\intercal + \Cbf \Bbf^\intercal + \Bbf \Cbf^\intercal  )$, and $(\Cbf \Cbf^\intercal +  \Ibf )^{-1}$ are symmetric.

First, let us simplify the equation as:
\begin{align}
    \Abf \Xbf \Bb  + \Xbf &= \Fbb,  
\end{align}
where $\Bb = ( n \Bbf \Bbf^\intercal + \Cbf \Bbf^\intercal + \Bbf \Cbf^\intercal  ) (  \Cbf \Cbf^\intercal +  \Ibf )^{-1} $  and  $\Fbb = -\Fbf (  \Cbf \Cbf^\intercal +  \Ibf )^{-1}$.
We perform eigen-decomposition on matrix $\Abf$ and $\Bb$, i.e.:
\begin{align}
    \Abf = \Ubf \Sbf \Ubf^{-1},
\end{align}
where $\Ubf$ is a matrix whose $i$-th column is the eigenvector $\uvec_i$ of $\Abf$, and $\Sbf$ is a diagonal matrix whose diagonal elements are the corresponding eigenvalues, $\Sbf_{ii} = \lambda_i$. Similarly, we also have:
\begin{align}
    \Bb = \Vbf \Tbf \Vbf^{-1},
\end{align}
where $\Vbf$ is a matrix whose $i$-th column is the eigenvector of $\Bb$, and $\Tbf$ is a diagonal matrix whose diagonal elements are the corresponding eigenvalues of $\Bb$. 

To make sure that we can apply the technique, we check the eigendecomposability of $\Abf$ and $\Bb$. Since $\Abf$ is symmetric, it is surely eigendecomposable. The matrix $\Bb$ may not be symmetric. However, both ${\bf \Bbar} = ( n \Bbf \Bbf^\intercal + \Cbf \Bbf^\intercal + \Bbf \Cbf^\intercal  )$ and ${\bf \Cbar} = (\Cbf \Cbf^\intercal +  \Ibf )^{-1}$ are symmetric. Based on  matrix similarity property, since $\Bb = {\bf \Bbar} {\bf \Cbar}$, the eigenvalues of $\Bb$ are the same as the eigenvalues of 
$
    {\bf \Cbar}^{\frac12} {\bf \Bbar} {\bf \Cbar} {\bf \Cbar}^{-\frac12} = {\bf \Cbar}^{\frac12} {\bf \Bbar}  {\bf \Cbar}^{\frac12}
$,
which is symmetric. Therefore, $\Bb$ is also eigendecomposable.

Applying the eigendecomposition technique, we have:
\begin{align}
    \Abf \Xbf \Bb  + \Xbf &= \Fbb \\
    \Ubf \Sbf \Ubf^{-1} \Xbf \Vbf \Tbf \Vbf^{-1}  + \Xbf &= \Fbb \\
    \Ubf \Sbf (\Ubf^{-1} \Xbf \Vbf) \Tbf \Vbf^{-1}  + \Xbf &= \Fbb.
\end{align}
Denote $\Xbf^* = \Ubf^{-1} \Xbf \Vbf$. We then have:
\begin{align}
    \Ubf \Sbf \Xbf^* \Tbf \Vbf^{-1}  + \Xbf &= \Fbb \\
    \Ubf^{-1} \Ubf \Sbf \Xbf^* \Tbf \Vbf^{-1} \Vbf  + \Ubf^{-1} \Xbf \Vbf  &= \Ubf^{-1}  \Fbb \Vbf  \\
    \Sbf \Xbf^* \Tbf  + \Xbf^*  &= \Ubf^{-1}  \Fbb \Vbf 
\end{align}

Let $\Gbf = \Ubf^{-1}  \Fbb \Vbf$. Since both $\Sbf$ and $\Tbf$ are diagonal matrices, we can solve for $\Xbf^\diamond$ easily by solving element-wise equations, i.e.:
\begin{align}
    \Xbf^\diamond_{i,j} (\Sbf_{i,i} \Tbf_{j,j} + 1) &= \Gbf_{i,j} \\
    \Xbf^\diamond_{i,j} &= \frac{\Gbf_{i,j}}{\Sbf_{i,i} \Tbf_{j,j} + 1}.
\end{align}
We can then easily recover $\Xbf$ from $\Xbf^\diamond$ by computing:
\begin{align}
    \Xbf = \Ubf \Xbf^\diamond \Vbf^{-1}.
\end{align}

When applying the decomposition technique above to the ADMM optimization, only the matrix $\Fbf$ changes in each iteration. All other matrices are fixed based on the form of the optimized performance metric. Therefore, we only perform the eigendecomposition once and store most of the required variables for the computation. This left us with just a few matrix multiplication operations that need to be computed for each ADMM iteration.

\subsection{Runtime Analysis}

For a batch of $m$ samples, all of the matrix variables in the ADMM formulations are $m \times m$ matrices. 
We run the ADMM algorithm for solving the inner optimization over $\Qbf$ in a fixed number of iterations (i.e., 100 iterations). In each iteration, we need to perform updates over the primal variables $\Qbf$, $\Zbf$, and $\Xbf$. 
In updating $\Qbf$, we perform a projection algorithm to the set $\Delta$. The runtime of the projection consists of sorting $m$-columns of $m$-items which costs $m^2 \log m$ in total. The iterative algorithm for finding the best $\pvec_k$ requires scanning the list, which costs $O(m)$ for each column, or $O(m^2)$ in total. The one-dimensional optimization for finding the optimal $\eta$ converges very quickly. We cap the number of iterations of finding $\eta$ to be at most 20 iterations. Hence, the total runtime of the projection algorithm is $O(m^2 \log m)$.
The algorithm for computing the prox function in $\Zbf$ updates costs the same as the projection algorithm. For solving the Sylvester equation, we need to perform eigendecomposition once, which costs $O(m^3)$. For every ADMM iterations, we only need to perform a few matrix multiplication operations, which costs $O(m^{2.5})$. Therefore, the total runtime complexity for solving the inner optimization over $\Qbf$ using our ADMM algorithm is $O(m^3)$.
